\documentclass[10pt,journal,twoside]{IEEEtran}
\usepackage{graphicx}
\usepackage{amsmath,bm}
\usepackage{tabularx,colortbl}
\usepackage{multirow}
\usepackage{booktabs}  
\usepackage{threeparttable} 

\usepackage[colorlinks]{hyperref}
\usepackage{cite}

\begin{document}
\title{%Just Noticeable Difference Estimation Based on A Deep-Represented Human Visual System
HVS-Inspired Signal Degradation Network for Just Noticeable Difference Estimation
}

\author{Jian~Jin,~\IEEEmembership{Member,~IEEE,}
        Yuan~Xue,
        Xingxing~Zhang,
        Lili~Meng,\\
        Yao~Zhao,~\IEEEmembership{Senior Member,~IEEE},
        and~Weisi~Lin,~\IEEEmembership{Fellow,~IEEE}% <-this % stops a space
\thanks{Copyright \copyright 20XX IEEE. Personal use of this material is permitted. However, permission to use this material for any other purposes must be obtained from the IEEE by sending an email to pubs-permissions@ieee.org. \emph{(Corresponding author: Weisi Lin.)}}

\thanks{J. Jin and W. Lin are with the School of Computer Science and Engineering, Nanyang Technological University, 639798, Singapore and also with Alibaba-NTU Singapore Joint Research Institute, Nanyang Technological University, 639798, Singapore. E-mail: jian.jin@ntu.edu.sg; wslin@ntu.edu.sg.}

\thanks{Y. Xue is with the School of Software, Fudan University, 200000, China. E-mail: 19212010045@fudan.edu.cn.}

\thanks{X. Zhang is with the Department of Computer Science and Technology, Tsinghua University, Beijing 100084, China. E-mail: xxzhang2020@mail.tsinghua.edu.cn.}

\thanks{L. Meng is with the School of Information Science and Engineering, Shandong Normal University, Jinan, 250014, China. E-mail: mengll\_83@hotmail.com.}

\thanks{Y. Zhao is with is with the Institute of Information Science, Beijing Jiao Tong University, 100044, China. E-mail: yzhao@bjtu.edu.cn.
}% <-this % stops an unwanted space

}

\markboth{Submit to IEEE Transactions on Cybernetics.}{Jin {\it \lowercase{et al.}}: HVS-Inspired Signal Degradation Network for Just Noticeable Difference Estimation}

\maketitle

\begin{abstract}

Significant improvement has been made on just noticeable difference (JND) modelling due to the development of deep neural networks, especially for the recently developed unsupervised-JND generation models. However, they have a major drawback that the generated JND is assessed in the real-world signal domain instead of in the perceptual domain in the human brain. There is an obvious difference when JND is assessed in such two domains since the visual signal in the real world is encoded before it is delivered into the brain with the human visual system (HVS). Hence, we propose an HVS-inspired signal degradation network for JND estimation. To achieve this, we carefully analyze the HVS perceptual process in JND subjective viewing to obtain relevant insights, and then design an HVS-inspired signal degradation (HVS-SD) network to represent the signal degradation in the HVS. On the one hand, the well learnt HVS-SD enables us to assess the JND in the perceptual domain. On the other hand, it provides more accurate prior information for better guiding JND generation. Additionally, considering the requirement that reasonable JND should not lead to visual attention shifting, a visual attention loss is proposed to control JND generation. Experimental results demonstrate that the proposed method achieves the SOTA performance for accurately estimating the redundancy of the HVS. Source code will be available at \textcolor[RGB]{202,12,22}{https://github.com/jianjin008/HVS-SD-JND}.
\end{abstract}

\IEEEpeerreviewmaketitle

\section{Introduction}
\label{sec:introduction}

\IEEEPARstart{J}{ust} noticeable difference (JND) \cite{chou1995perceptually,yang2005motion,liu2010just,wu2013just,wu2017enhanced,jia2006estimating,bae2013novel,bae2014novel,niu2013visual,hadizadeh2017saliency,zeng2019visual,bae2016dct,jin2016statistical,wang2016mcl,wang2017videoset,liu2018jnd,lin2022large,liu2019deep,zhang2021deep,tian2021perceptual,wu2020unsupervised,jin2021just,jin2022full} as a significant metric reflects the perceptual redundancy in visual signals for the human visual system (HVS) \cite{hall1977nonlinear}, which has been widely used in multimedia signal processing, such as perceptual image and video compression \cite{wu2013perceptual,kim2015hevc,zhang2019divisively,zhou2020just,nami2022bl}, watermarking \cite{jia2020rihoop,cheng2001additive}, image and video quality enhancement \cite{su2020joint,gu2015analysis}, and so on \cite{shen2013depth,shen2014exposure,zhu2013learning}. Generally, JND refers to minimum visual signal changes that the HVS is able to perceive, which can be regarded as a set of thresholds. The HVS has been studied for decades. However, as the limitation of our knowledge and technologies, the HVS cannot be fully understood so far. Hence, how to accurately estimate JND and build JND models that are highly consistent with the HVS remain an open question. Commonly, JND models can be divided into two categories, i.e., HVS-inspired JND models \cite{chou1995perceptually,yang2005motion,liu2010just,wu2013just,wu2017enhanced,jia2006estimating,bae2013novel,bae2014novel,niu2013visual,hadizadeh2017saliency,zeng2019visual,bae2016dct} and learning-based JND models \cite{liu2019deep,zhang2021deep,tian2021perceptual,jin2016statistical,wang2016mcl,wang2017videoset,liu2018jnd,lin2022large,wu2020unsupervised,jin2021just,jin2022full}. 

HVS-inspired JND models are commonly designed based on the limited knowledge about the HVS that we can understand, e.g., luminance adaption (LA) \cite{bae2013novel} and contrast masking (CM) \cite{bae2014novel} effects on the HVS, visual attention (VA) \cite{niu2013visual,hadizadeh2017saliency,zeng2019visual} and visual sensitivity (VS) \cite{cong2022global,sun2022tensorial} effects on the JND, different types of retina cells and their spatial distribution (e.g, fovea \cite{bae2016dct}), and so on. Many visual maskings (termed as handcrafted features) are developed based on them and used for JND modelling, which achieves good performance. However, there remains a gap between the actual redundancy in the HVS and the redundancy estimated by the HVS-inspired JND model, as much of the knowledge about the HVS is still missing. 

Recently, with the development of deep neural networks, massive labeled datasets as well as high-performance GPU hardware, breakthroughs have been made in learning-based JND modelling. Generally, learning-based JND models can be further divided into two sub-categories, i.e., fully supervised learning-based JND models \cite{liu2019deep,zhang2021deep,tian2021perceptual,jin2016statistical,wang2016mcl,wang2017videoset,liu2018jnd,lin2022large} and weakly/unsupervised learning-based JND models \cite{wu2020unsupervised,jin2021just,jin2022full}. The former ones highly rely on labeled JND datasets \cite{jin2016statistical,wang2016mcl,wang2017videoset,liu2018jnd,lin2022large}. However, the existing JND datasets only take the compression-caused distortion into account so that the developed JND models can be used for perceptual coding. Hence, the former ones cannot estimate a generalized JND due to ignoring the other kinds of distortions during JND modelling. So far, there are at least 42 distortion types. It’s infeasible to build the JND datasets to cover all kinds of distortion types considering the huge manpower and time cost of JND data labeling. In view of this, many weakly/unsupervised learning-based JND models have been developed and made breakthroughs. They generate the JND as the noise via generative networks (e.g., auto-encoder networks \cite{wu2020unsupervised,jin2021just,jin2022full}) and assess its reasonability via image quality assessment (IQA) metrics until the generated JND does not cause image quality degradation. Such kind of models have two advantages. 
\begin{figure}[htbp]
	\begin{center}
		\noindent
		\includegraphics[width = 3.4 in]{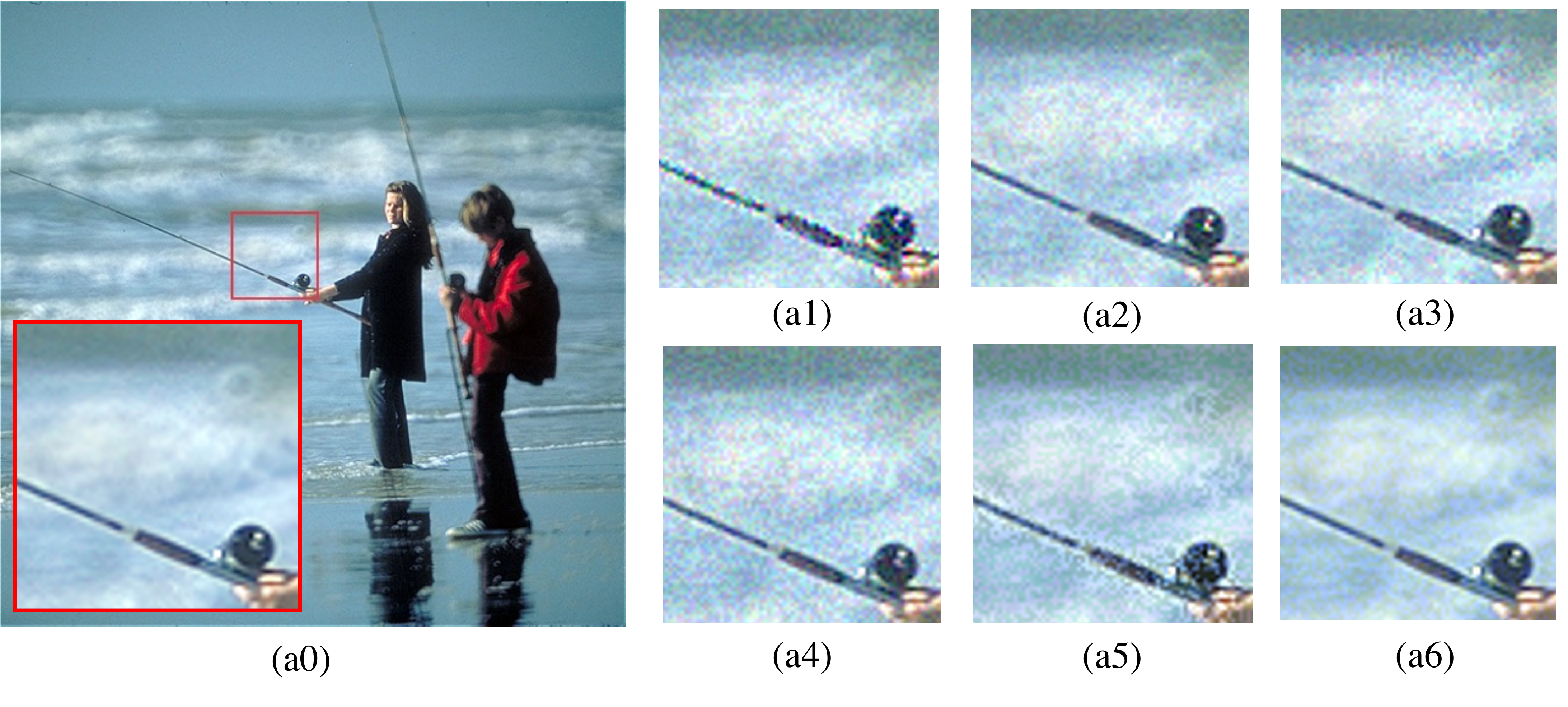}
		\caption{Details comparison of JND distorted images. Due to the limit of space, only the full original image (a0) is shown in this figure. The red block region of (a0) is enlarged. (a1)-(a6) are generated by injecting the JND, estimated with previous SOTA JND methods \cite{liu2010just,wu2013just,wu2017enhanced,wu2020unsupervised,jin2022full} and our proposed method, into original image. Although they have the same PSNR (26.06 dB), the injected JND noise in our JND distorted image (a6) can hardly be perceived compared with the others, as the closest image to the original one. 
		}\label{intro_image}
	\end{center}
\end{figure}
Firstly, IQA metrics are used to replace the JND subjective viewing test to assess the reasonability of the generated JND, which makes them doesn’t rely on the labeled JND data. Secondly, during JND generation, the generative networks can simulate any kind of distortion, which makes all kinds of distortions been considered during training and enables them to estimate a more generalized JND. However, there are a major disadvantage. That is, the reasonability of the generated JND should be perceived by the subjects in perceptual domain instead of assessing with the IQAs in signal domain. Besides, they highly rely on handcrafted features for guiding JND generation, which are designed based on partial characteristics of the HVS that we can understand and limit their accuracy. The disadvantages above lead to the gap between the generated JND and the actual redundancy of the HVS.

In this work, we propose an HVS-inspired signal degradation network for JND estimation. To achieve this, we carefully analyze the HVS perceptual process in JND subjective viewing to obtain relevant insights. Firstly, the perceptual difference in our brain's visual cortex is different from the signal-fidelity difference in the real world due to the signal degradation during the HVS perceptual process. That is, visual signal has been encoded by the optic nerve before it's delivered to the brain's visual cortex. Secondly, in JND subjective viewing tests, as the main function of the HVS is to perceive the detailed difference between the original image and its associated JND distorted image, signal degradation can be represented with a lossy image codec. Thirdly, unreasonable JND estimation will be easily perceived by the HVS, which leads to visual attention focusing on incorrect JND regions and results in visual attention shifting. 

Based on the first two obtained insights above, we propose an HVS-inspired signal degradation network, termed the HVS-SD, which can be regarded as a learnt lossy image codec. With the proposed HVS-SD, we can predict perceptual images in our brain in JND subjective viewing so that the difference between the original image and its associated JND-distorted image can be assessed in the perceptual domain. Besides, compared with the handcrafted features used in the weakly/unsupervised learning-based JND models, more accurate prior information, such as visual attention and sensitivity, are obtained from the HVS-SD via Grad-CAM technique \cite{selvaraju2017grad}, which can better guide JND generation. Additionally, we propose a visual attention loss based on the third obtained insight above to further control the JND generation and guarantee that the generated JND will not cause attention shifting of the HVS. All the advantages above make our proposed method achieve the SOTA result on JND estimation, as shown in Fig. \ref{intro_image}. The contributions of this work are summarized as follows:

\begin{itemize}
    \item We analyze HVS perceptual process in JND subjective viewing and represent the signal degradation in the HVS with a proposed HVS-SD network. The well-learnt HVS-SD enables the generated JND to be assessed in the perceptual domain (instead of the real-world signal one). 
    
    \item We propose a visual attention loss to further control JND generation. Compared with visual attention used as weights or features in the existing JND models, to the best of our knowledge, this is the first work that explicitly utilizes visual attention as a loss during JND estimation.
    
    \item We evaluate our method against previous SOTA ones on the CSIQ \cite{larson2010categorical}. The experimental results well demonstrate that both the proposed HVS-SD and visual attention loss significantly improve the accuracy of JND estimation.
\end{itemize}

\section{Related Works}
\subsection{JND Models}

\subsubsection{HVS-inspired JND models}
As JND is a result of the perceptual phenomenon of the HVS and reflects the redundancy of the HVS, the characteristics of the HVS are considered in most of the initial JND models \cite{chou1995perceptually,yang2005motion,liu2010just,wu2013just,wu2017enhanced,jia2006estimating,bae2013novel,bae2014novel,niu2013visual,hadizadeh2017saliency,zeng2019visual,bae2016dct}. Chou \emph{et al.} \cite{chou1995perceptually} first combined the luminance adaptation (LA) and contrast masking (CM) and proposed a spatial-domain JND model. To accurately characterize the LA and CM, especially for their overlaps, Yang \emph{et al.} \cite{yang2005motion} proposed a generalized spatial JND model by introducing a nonlinear additivity model for masking effects (NAMM). Meanwhile, the contrast sensitivity function (CSF) which reflects the characteristics of HVS in the spatial frequency domain was introduced in modelling the JND in the frequency domain by Jia \emph{et al.} \cite{jia2006estimating}. The CFS, LA, and CM were combined by Bae \emph{et al.} \cite{bae2013novel,bae2014novel} and further proposed a new sub-band JND profile \cite{bae2016dct}. After that, edge masking (EM) and texture masking (TM) were first proposed in \cite{liu2010just} and used for enhancing the JND model. Then, the free-energy principle was first introduced in JND modelling by Wu \emph{et al.} \cite{wu2013just}, which achieves a comprehensive evaluation. Moreover, they utilized the pattern complexity (PC) \cite{wu2017enhanced} of visual content (regarded as the visual sensitivity of the HVS) to further improve the accuracy of the JND model. Meanwhile, the visual attention (VA) was also utilized as weights in some JND works \cite{niu2013visual,hadizadeh2017saliency,zeng2019visual} to customize different masking during the JND modelling.     

\begin{figure}[htbp]
	\begin{center}
		\noindent
		\includegraphics[width = 3.3 in]{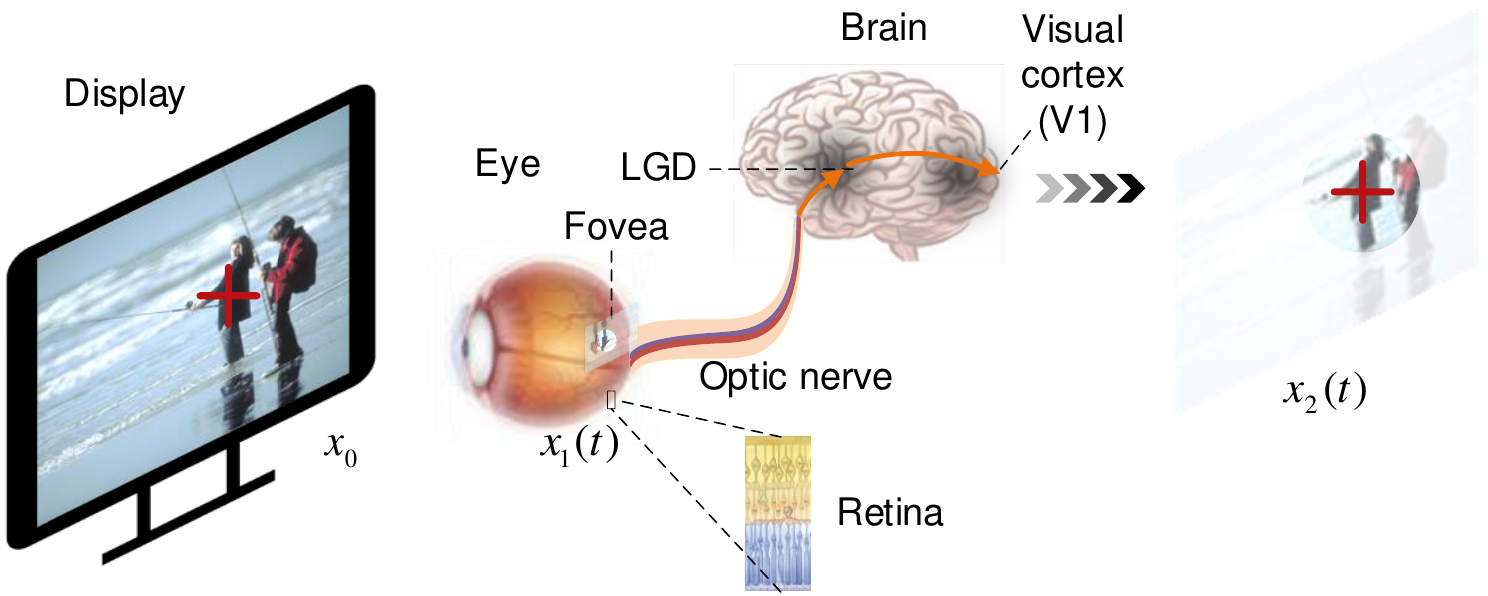}
		\caption{Illustration of the HVS perceptual process with focus on a certain region (red cross) of one displayed image $x_0$ at a certain moment $t$. As the existence of the fovea, the focused region closed to the red cross is commonly captured with higher resolution and forms the reflection $x_1(t)$. Then, the information of $x_1(t)$ is delivered to the human brain via optic nerve and forms perceptual image $x_2(t)$ in the visual cortex.}\label{HVS}
	\end{center}
\end{figure}

\subsubsection{Learning-based JND models}
Considering deep learning achieved significant success in lots of visual problems, lots of JND datasets and learning-based JND models were proposed recently \cite{liu2019deep,zhang2021deep,tian2021perceptual,jin2016statistical,wang2016mcl,wang2017videoset,liu2018jnd,lin2022large,wu2020unsupervised,jin2021just,jin2022full}. Jin \emph{et al.} \cite{jin2016statistical} proposed the first JND image dataset, named MCL-JCI dataset, where the picture-wise JND concept was first proposed. Then, a large-scale picture-wise JND dataset was proposed by Lin \emph{et al.} \cite{lin2022large} via a crowdsourced subjective assessment method. After that, a video JND dataset, named VideoSet, was built by Wang \emph{et al.} \cite{wang2017videoset}, where they proposed the satisfied-user-ratio (SUR) defined the video-wise JND. Then, Liu \emph{et al.} \cite{liu2019deep} used deep learning technique to predict picture-wise JND points for compressed images based on the MCL-JCI dataset, where the picture-wise JND prediction was formulated as a classification problem. Afterward, a video-wise JND and SUR model were learned by Zhang \emph{et al.} \cite{zhang2021deep} by taking the temporal and spatial information into account. After that, Wu \emph{et al.} \cite{wu2020unsupervised} proposed an unsurprising/weakly-supervising learning method for JND modelling, where the JND is generated as noise via a generative network and measured with an IQA metric (i.e., SSIM). Meanwhile, Jin \emph{et al.} \cite{jin2021just} proposed the first JND for the deep machine vision, where the performance of the machine vision task is used to supervise the JND generation, which is the first exploration of JND on deep machine vision. Then, considering that most of the existing JND models only take the stimulus of the main channel into account, Jin \emph{et al.} \cite{jin2022full} proposed the first full RGB-JND model, where the stimuli of the whole color space were taken into account during JND modelling. Besides, they also utilized the handcrafted features (e.g., PC and VA) for guiding JND generation. The generated JND was measured by their proposed adaptive IQA combination (AIC) module, and achieved the SOTA performance. However, all these unsurprising/weakly-supervising learning methods \cite{wu2020unsupervised,jin2021just,jin2022full} above assessed the generated JND in the signal domain rather than the perceptual domain. 

\subsection{HVS Perceptual Process}
\label{pp-HVS}
The perceptual characteristics of the HVS have been widely used in solving visual signal processing and computer vision problems, e.g., image and video compression \cite{wu2013perceptual,kim2015hevc,zhou2020just,nami2022bl}, JND modelling \cite{chou1995perceptually,yang2005motion,liu2010just,wu2013just,wu2017enhanced,jia2006estimating,bae2013novel,bae2014novel,niu2013visual,hadizadeh2017saliency,zeng2019visual,bae2016dct}, image and video quality assessment \cite{galkandage2020full}, image and video enhancement \cite{yu2018perceptually}, and so on. The HVS is composed of the eyes, optic nerve, lateral geniculate nucleus (LGN) \cite{o2002attention}, and visual cortex/center (e.g., primary visual cortex V1 \cite{tootell1998functional}, and higher-order visual cortex V2, V3, V4, ... \cite{ekstrom2008bottom}). Here, an example of the HVS perceptual process is reviewed when it focuses on a certain region of one displayed image at a certain moment, as shown in Fig. \ref{HVS}. For simplicity, one eye of two is illustrated in this figure. 

Assume that the eye focuses on the red cross region of the displayed image $x_0$ at moment $t$ at first. As the existence of the fovea, the light of the focused region (closed to the red cross) is captured with higher resolution in the outermost layer of the retina (composed of around 130 million photoreceptor cells, e.g., rods and cones \cite{jonas1992human}), while the light of rest of regions is captured with lower resolution. Then, the reflection $x_1(t)$ appears in the retina. Then, the light signal of $x_1(t)$ is transformed into a neural signal and delivered to our brain via the optic nerve (composed of around 1 million nerve fibers \cite{zrenner2002will}). After the LGN and primary visual cortex (e.g., V1) processing, the corresponding perceptual image $x_2(t)$ is reconstructed in our brain, in which the resolution of the focused region remains higher and the rest of the regions are blurred. Compared with $x_0$ and $x_1(t)$, $x_2(t)$ is what we actually perceive at moment $t$. Considering that the number of photoreceptor cells is 130 times larger than that of the optic nerve fibers, it’s commonly believed that there is a lossy compression before $x_1(t)$ is delivered to the optic nerve \cite{gao2021digital}, which results in the so-called signal degradation. Hence, the actual perceived image $x_2(t)$ at a certain moment $t$ is slightly different from the reflection image $x_1(t)$ and largely different from the displayed image $x_0$. After that, $x_2(t)$ is used for high level visual understanding, such as motion/depth perception \cite{burr1994selective}, form recognition \cite{taylor2022representation}, and so on, in the higher-order visual cortex.

\section{Proposed Method}
% In this section, we firstly revisit the HVS perceptual process during the JND subjective viewing test and obtain some important observations. Then, we simulate this process with a proposed HVS-SD. After that, a new JND estimation model based on the proposed HVS-SD is developed, where more accurate prior information will be used for guiding JND generation, and visual attention is first developed as a loss for JND modelling. 

In this section, we firstly analyze the HVS perceptual process in the JND subjective viewing test to obtain some important insights. Based on the insights, we propose an HVS-SD network to represent the signal degradation that existed in the HVS perceptual process. After that, a new JND estimation model based on the proposed HVS-SD is developed, where the proposed HVS-SD enables the generated JND to be assessed in the perceptual domain. Meanwhile, more accurate prior information is obtained from the HVS-SD and used for guiding JND generation. Finally, a visual attention loss is developed to better control the JND generation. 

\subsection{Analysis of JND Subjective Viewing Tests}
\label{SVT}

%JND subjective viewing test aims to assess if there are differences that can be perceived by the subjects between the original image and its associated JND-distorted image.

\begin{figure}[htbp]
	\begin{center}
		\noindent
		\includegraphics[width = 3.3 in]{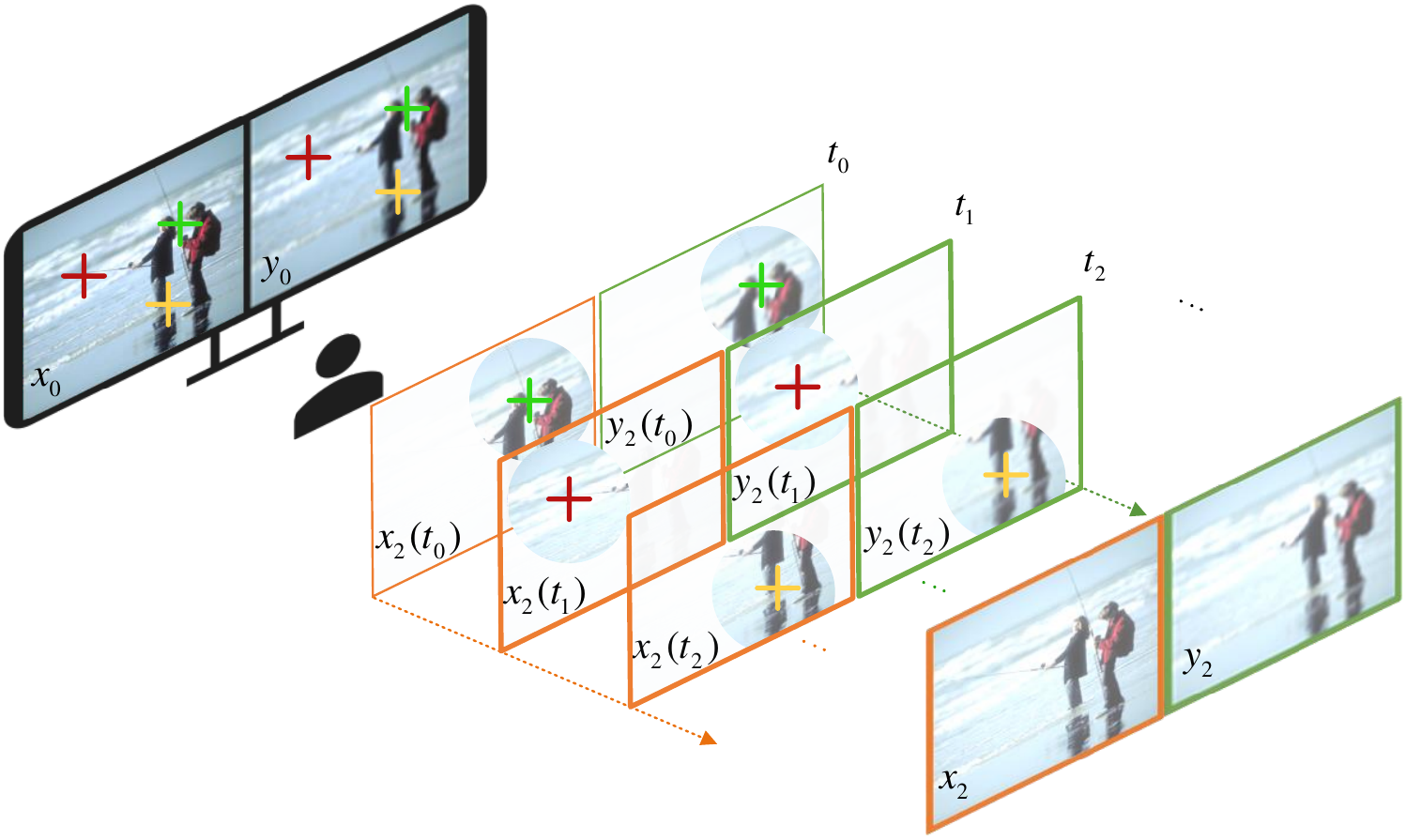}
		\caption{Illustration of the HVS perceptual process during JND subjective viewing tests at different moments. Focus regions at different moments are represented with crosses of different colors (e.g., \textcolor{green}{green} crosses at $t_0$, \textcolor{red}{red} crosses at $t_1$, \textcolor{yellow}{yellow} crosses at $t_2$, and so on.)}\label{JNDtest}
	\end{center}
\end{figure}

To assess the reasonability of JND, the JND subjective viewing test is commonly carried out by perceiving if there is a difference between the original image and its associated JND-distorted image (obtained by injecting the JND into the original image). It should be noticed that the difference mentioned above mainly refers to the low-level visual form difference (e.g., color, details, and so on). High-level visual understanding differences (e.g., motion, depth, and so on) are not involved in the JND subjective viewing test. Its experimental environment commonly refers to the guidance of ITU-R BT.500-11 standard \cite{bt2002methodology}. For each turn of an experiment individual, as shown in Fig. \ref{JNDtest}, the original image $x_0$ and its associated JND-distorted image $y_0$ are exhibited on the left and right sides of the same display. 
During the test, the subjects have enough time to go through the images and compare their details from point to point according to their preferences. Fig \ref{JNDtest} shows an example of the focus shift from a subject at different moments (e.g., moment $t_0$, $t_1$, $t_2$, ...) when the subject compares these two images by point-to-point. According to the review of the HVS perceptual process in Sec. \ref{pp-HVS}, only the content closed to the focus is able to be perceived as higher resolution (e.g., $x_2(t_0)$, $y_2(t_0)$, $x_2(t_1)$, $y_2(t_1)$, $x_2(t_2)$, $y_2(t_2)$, ...), and used for perceptual comparison. Therefore, the actual perceptual comparisons are mainly performed on the focus regions at different moments. As subjects have enough time to go through these two images, the perceptual comparisons in the JND subjective viewing test can be approximately equivalent to the comparisons on the accumulations of all the focus regions from two images, i.e, the comparison between $x_2=x_2(t_0+t_1+t_2+…)$ and $y_2= y_2(t_0+t_1+t_2+…)$. Due to the signal degradation of the HVS perceptual process as reviewed in Sec. \ref{pp-HVS}, $x_2$ (or $y_2$) is different from $x_0$ (or $y_0$). Hence, the perceptual difference (the difference between $x_2$ and $y_2$) is different from the signal-fidelity difference (the difference between $x_0$ and $y_0$). 

Besides, $x_2$ and $y_2$ are the accumulations of the focus regions, which are with high signal fidelity. Therefore, $x_2$ and $y_2$ can be regarded as the high-quality decompressed images of $x_0$ and $y_0$. The signal degradation of the HVS perceptual process in the JND subjective viewing can be represented with a lossy image codec (denoted by $\mathcal{C}(\cdot)$), which can decode high-quality images. Then, we have 
\begin{equation}
\label{codec}
\left\{
\begin{aligned}
& x_2=\mathcal{C}(x_0)\\
& y_2=\mathcal{C}(y_0)
\end{aligned}.
\right.
\end{equation}

Additionally, a fact in \cite{dong2011selective} shows that the unexpected distortion may cause the visual attention shift. Similarly, once the unreasonable JND is perceived by the HVS from the JND-distorted image, more attention will be assigned to such regions. This will cause the visual attention of the JND-distorted image to be different from that of the original image. In other words, unreasonable JND is just like the unexpected distortion, which may cause visual attention shift as well.

\begin{figure}[htbp]
	\begin{center}
		\noindent
		\includegraphics[width = 3.5 in]{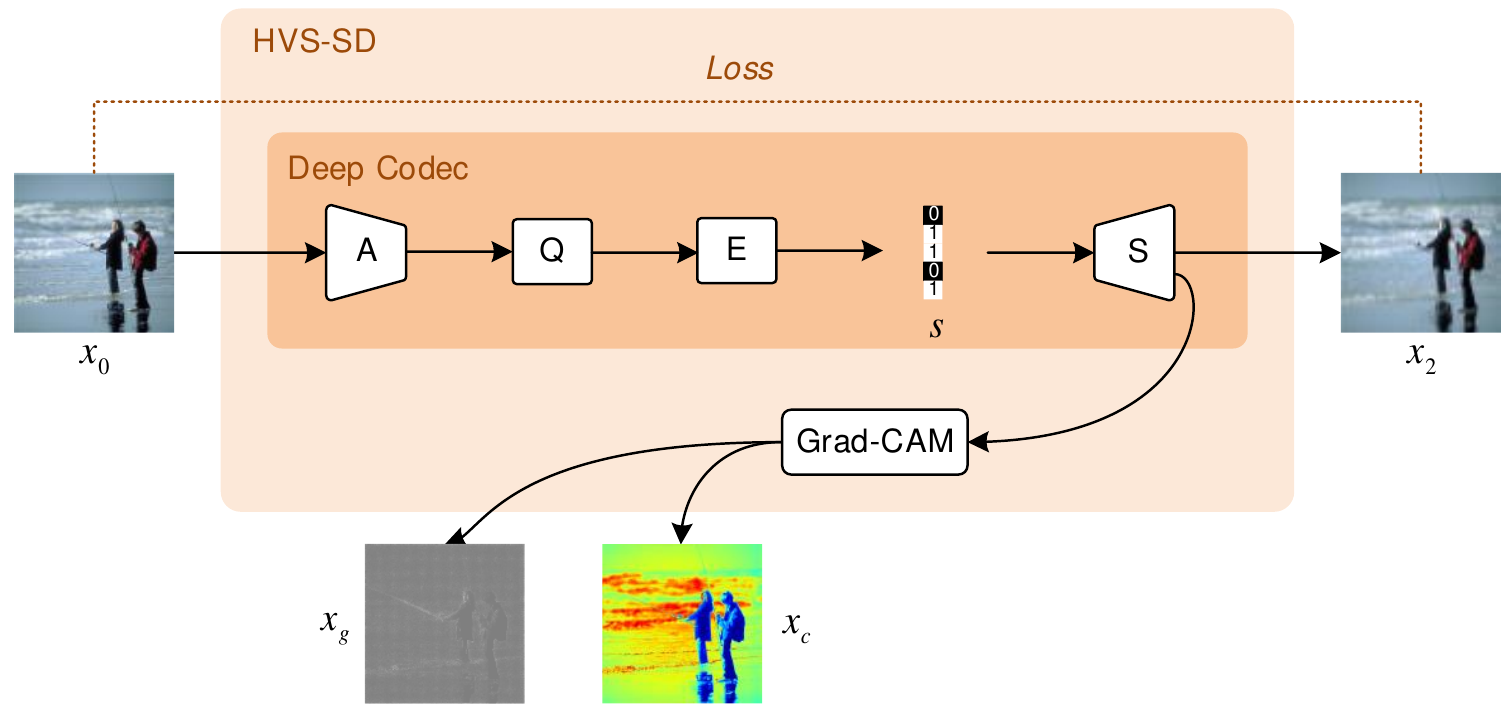}
		\caption{The framework of the proposed HVS-SD network.}\label{fig-HVS-SD}
	\end{center}
\end{figure}
\begin{table}[htbp]
  \centering
  \caption{\scshape Detailed Structure of The Deep Codec.}
    \begin{tabular}{lcc}
    \toprule
    Layer & Input size & Output size \\
    \midrule
    Conv1 & w,h,3 & w/2,h/2,128 \\
    GDN1  & w/2,h/2,128 & w/2,h/2,128 \\
    Conv2 & w/2,h/2,128 & w/4,h/4,128 \\
    GDN2  & w/4,h/4,128 & w/4,h/4,128 \\
    Conv3 & w/4,h/4,128 & w/8,h/8,128 \\
    Deconv1 & w/8,h/8,128 & w/4,h/4,128  \\
    iGDN1 & w/4,h/4,128  & w/4,h/4,128  \\
    Deconv2 & w/4,h/4,128 & w/2,h/2,128 \\
    iGDN2 & w/2,h/2,128 & w/2,h/2,128 \\
    Deconv3 & w/2,h/2,128 & w,h,3 \\
    \bottomrule
    \end{tabular}%
  \label{tab:deepcodec}%
  \vspace{-0.6cm}
\end{table}%
So far, for the existing learning-based JND models, the generated JND is directly assessed in the signal domain (between $x_0$ and $y_0$) instead of being assessed in the perceptual domain (between $x_2$ and $y_2$), which is unreasonable. Besides, there is hardly any work on representing the signal degradation of the HVS perceptual process for JND modelling via deep learning techniques. Moreover, although visual attention is closely connected with the JND, no visual attention loss is developed for the JND modelling in the existing learning-based JND models.

Based on the insights above, we represent the signal degradation of the HVS perceptual process in the JND subjective viewing test with the proposed HVS-SD network in Sec. \ref{HVS-SD}. On the one hand, the well-learnt HVS-SD enables JND to be assessed in the perceptual domain. On the other hand, it provides more accurate prior information (e.g., visual attention and sensitivity) and better guide for the JND generation. Besides, a new framework for the JND generation is developed in Sec. \ref{HVS-SD-JND}, where the visual attention loss is first proposed for assessing the generated JND. 

\begin{figure*}[htbp]
    \begin{center}
        \noindent
        \includegraphics[width=17cm]{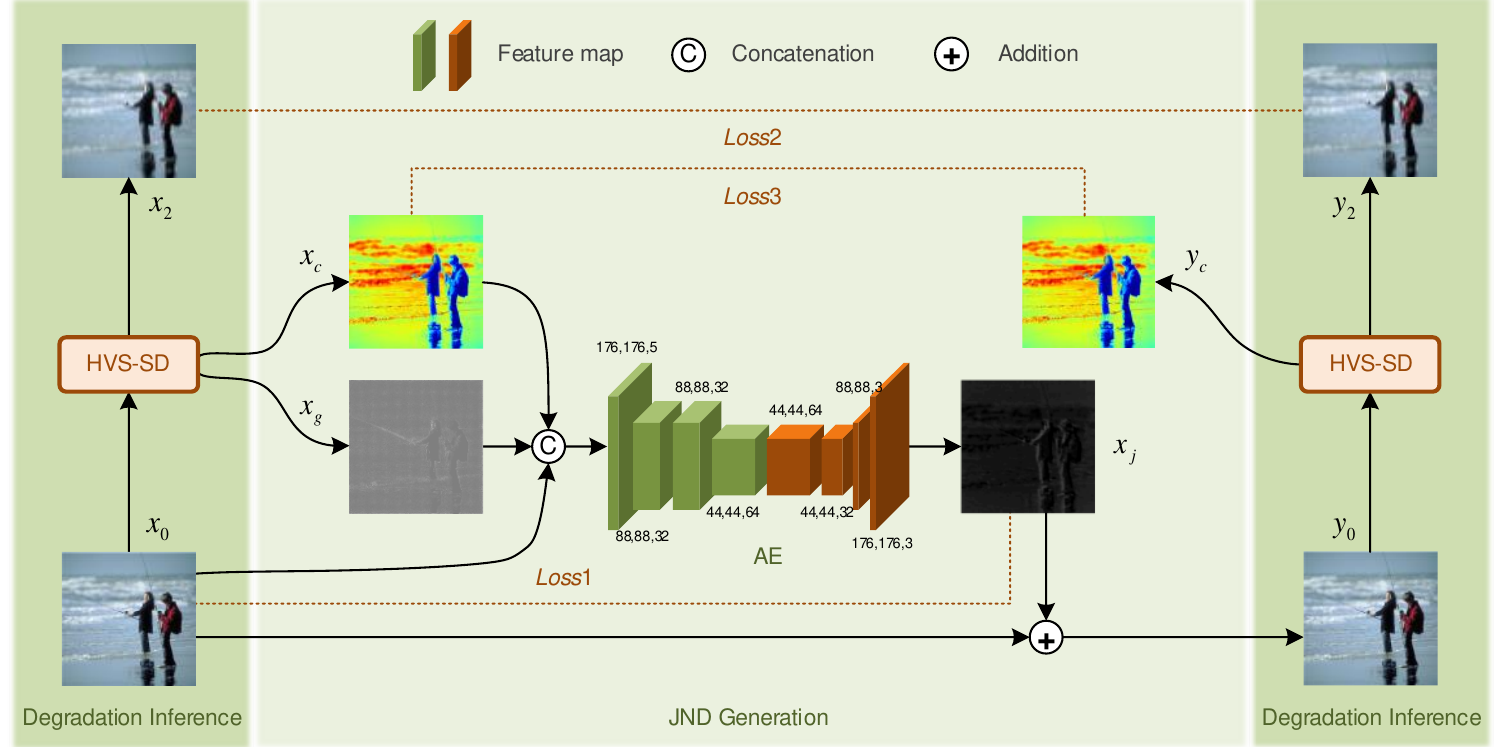}
        \caption{The framework of the proposed HVS-SD-JND. The original image $x_0$ is the input of the HVS-SD-JND. It's firstly fed into the well-trained HVS-SD network to predict its associated degraded image $x_2$. Meanwhile, the guided-CAM map $x_g$ and CAM map $x_c$ are obtained. After concatenated with the original image $x_0$, $x_0$ together with $x_g$ and $x_c$ are fed into the auto-encoder network to generate the JND $x_j$. After injecting $x_j$ into $x_0$ by applying the addition operation, JND distorted image $y_0$ is obtained. Then, its degraded image $y_2$ is inferred by feeding $y_0$ into the HVS-SD network. The HVS-SD networks used for inferring $x_2$ and $y_2$ share the same weights.   }
    \label{FW}
    \end{center}
\end{figure*}

\subsection{HVS-SD}
\label{HVS-SD}
An HVS-inspired signal degradation (HVS-SD) network is developed here to represent the signal degradation of the HVS perceptual process in the subjective viewing test. HVS-SD is composed of a typical deep codec \cite{balle2016end} and a Grad-CAM module \cite{selvaraju2017grad}. The deep codec is made up of four main components, namely analysis transform (A), quantizer (Q), entropy coder (E), and synthesis transform (S) components. The details of HVS-SD are shown in Fig. \ref{fig-HVS-SD}. Original image $x_0$ is firstly fed into the analysis transform. After that, the transformed latent vectors are quantized into entries. Then, entropy coding is applied to compress the entry vectors into stream $s$. After $s$ is delivered to the decoder sides, synthesis transform is applied and the decoded image $x_2$ is obtained. Meanwhile, with the Grad-CAM module, the guided-CAM map $x_g$ and the CAM map $x_c$ are obtained as well, which will be used for guiding the JND generation. Similarly, $y_2$ can be obtained by feeding $y_0$ into the HVS-SD. More details will be presented in Sec. \ref{HVS-SD-JND}.

Here, deep codec directs lossy compression to represent the signal degeneration, namely the aforementioned $\mathcal{C}(\cdot)$. To make sure that the perceived images $x_2$ and $y_2$ maintain high fidelity, we use a large amount of data to train deep codec $\mathcal{C}(\cdot)$ and maintain the decompressed images around 40dB PSNR. Decompressed images with 40dB PSNR are commonly considered of high fidelity, where the distortion can hardly be perceived \cite{abu2019crypto}. The detailed network structure of $\mathcal{C}(\cdot)$ refer to TABLE \ref{tab:deepcodec}, which is mainly composed of convolution/deconvolution and GDN/iGND layers. $w$ and $h$ denote the width and height of input, respectively. 

Besides, the MSE between the original image and its associated JND distorted image is used to calculate the CAM and guided-CAM in this work. Hence, the generated guided-CAM map and the CAM map can be regarded as the visual sensitivity and attention of the proposed HVS-SD when it maintains high fidelity image reconstruction. Compared with the handcrafted designed visual sensitivity and attention (i.e., PC and VA) maps in \cite{jin2022full}, our guided-CAM and CAM maps generated from the well-learnt codec are more accurate due to large prior information obtained from the large training data. Therefore, they are more suitable for guiding the JND generation. The guide-CAM map $x_g$ and CAM map $x_c$ can be calculated as
\begin{equation}
\label{xgxc}
(x_g, x_c) = \mathcal{G}(x_0, \mathcal{C}(\cdot)),
\end{equation}
%
% \begin{equation}
% \label{CAM}
% x_{c}=\operatorname{ReLU}\left(\sum_{k} \alpha_{k}^{M S E} A^{k}\right)
% \end{equation}
%
% \begin{equation}
% \label{Guided}
% x_{guided}=\frac{\partial y^{M S E}}{\partial A^{\text {image}}}
% \end{equation}
%
% \begin{equation}
% \label{Weight}
% \alpha_{k}^{M S E}=\frac{1}{W H} \sum_{i} \sum_{j} \frac{\partial y^{M S E}}{\partial A_{i j}^{k}}
% \end{equation}
%
% \begin{equation}
% \label{GCAM}
% x_{g} = x_{c}\cdot x_{guided}
% \end{equation}
%
where $\mathcal{G}(\cdot, \cdot)$ denotes the Grad-CAM technique.

\subsection{HVS-SD-JND}
\label{HVS-SD-JND}
In this subsection, a new framework for JND estimation is developed based on the HVS-SD, termed HVS-SD-JND. Besides, a visual attention loss is first proposed for assessing the generated JND. 

\subsubsection{Framework}
The framework of the HVS-SD-JND is made up of three main parts, namely the degradation inference parts of the original image and its associated JND-distorted image, as well as the JND generation part. The HVS-SD networks in two degradation inference parts share the same weights, which have been well-trained offline. The auto-encoder (AE) network in the JND generation part is to be trained in this work.  

As shown in Fig. \ref{FW}, $x_0$ as the input is firstly fed into the well-trained HVS-SD network, and degraded image $x_2$ is inferred. Meanwhile, the attention map $x_c$ and sensitivity map $x_g$ of image $x_0$ are obtained. After concatenated with the $x_0$, $x_c$, $x_g$, and $x_0$ are fed into the AE to generate the JND $x_j$. Then, the generated JND $x_j$ is injected into its associated original image $x_0$ by performing an addition operation, and JND distorted image $y_0$ is obtained. Then, $y_0$ is fed into the HVS-SD network and its perceptual image $y_2$ is obtained. Meanwhile, its attention map $y_c$ is obtained. It should be mentioned that the AE used here has the same structure as that used in \cite{jin2022full}. The detailed structure of AE has been shown in this figure.

\subsubsection{Loss Function}
To better control the magnitude of the generated JND in different content images, the magnitude loss $Loss1$ is inherited from \cite{selvaraju2017grad}, which is formulated as 
\begin{equation}
\label{loss1}
Loss1 = \ln (G^2 + x_j^2 + t_0) - \ln (2 \cdot G \cdot |x_j| + t_0),
\end{equation}
where $|\cdot|$ is the absolute operation and $t_0$ is a constant to avoid the denominator being zero. $G$ denotes the gradient of original image $x_0$, where $G = \sqrt{g_{0}(x_0)^2 + g_{1}(x_0)^2}$. $g_0(x_0)$ and $g_1(x_0)$ denotes the vertical and horizontal gradient of original image $x_0$.  

To assess the reasonability of the generated JND, the adaptive IQA combination (AIC) loss $Loss2$ is also inherited from \cite{selvaraju2017grad}. However, unlike the JND-distorted image being assessed in the signal domain, we assess its quality in the perceptual domain. That is, $x_2$ and $y_2$ are assessed with the $Loss2$ in this work instead of the $x_0$ and $y_0$ being assessed in \cite{selvaraju2017grad}. Then, $Loss2$ is formulated as
\begin{equation}
\label{loss2}
Loss2 = AIC(x_2,y_2).
\end{equation}

To make sure that the generated JND is reasonable and will not lead to visual attention shift, we developed visual attention loss $Loss3$ here, which is represented with the MSE between $x_c$ and $y_c$. Then, $Loss3$ is formulated as
\begin{equation}
\label{loss3}
Loss3 = MSE(x_c, y_c).
\end{equation}

Finally, the total loss $\mathcal{L}$ of the HVS-SD-JND can be formulated as
\begin{equation}
\label{Loss}
\mathcal{L} = \alpha \cdot Loss1 + \beta \cdot Loss2 + \gamma \cdot Loss3,
\end{equation}
where $\alpha$, $\beta$, and $\gamma$ are three parameters to balance their associated items in the formulation. The settings of such parameters will be presented in Sec. \ref{Exp}. 

\begin{figure*}[htbp]
    \begin{center}
        \noindent
        \includegraphics[width=16cm]{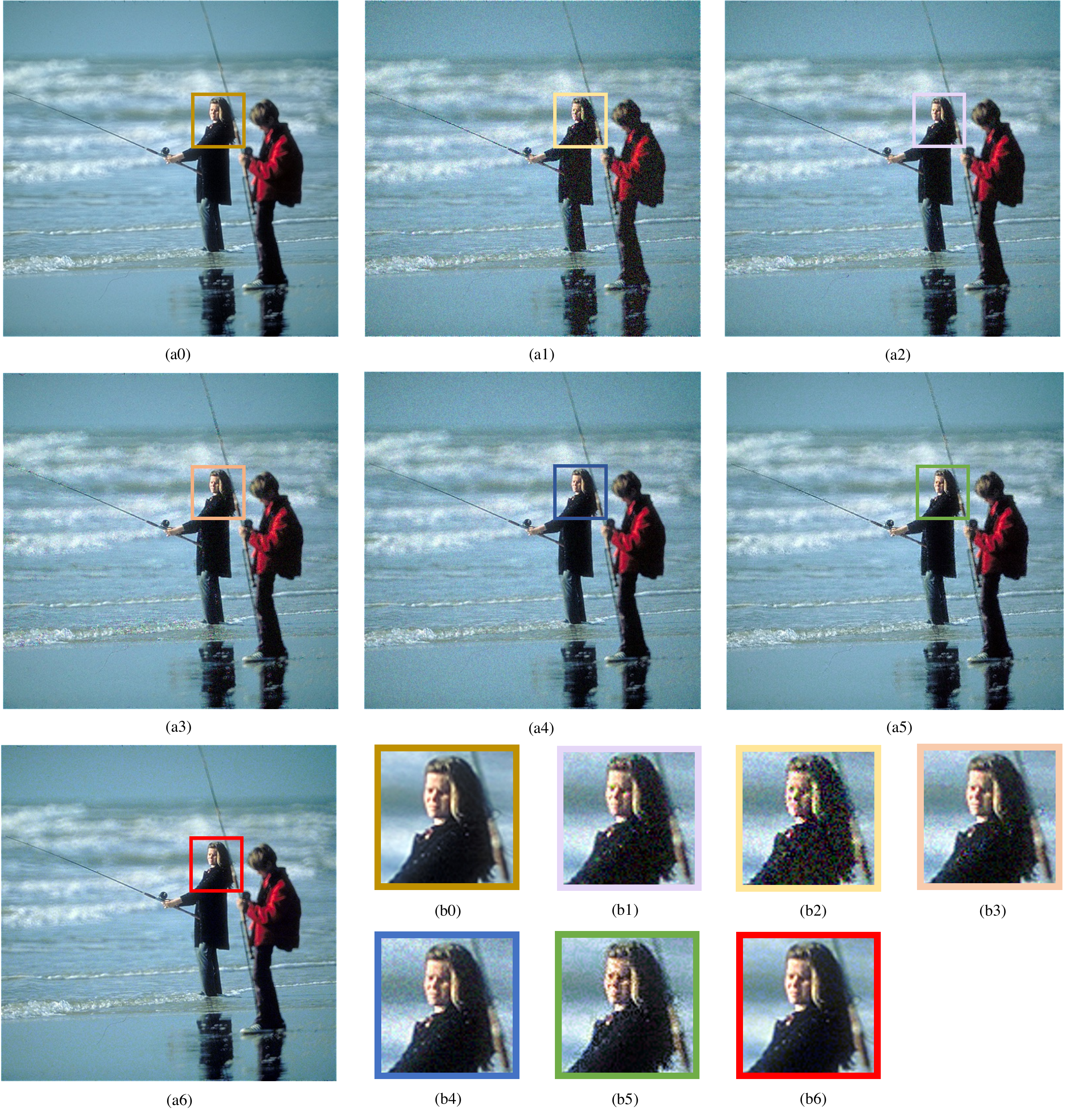}
        \caption{Comparison of images distorted by different JND models. (a0) is the original images. (a1), (a2), (a3), (a4), (a5) and (a6)are the images distorted by the JND model of Liu2010 \cite{liu2010just}, Wu2013 \cite{wu2013just}, Wu2017 \cite{wu2017enhanced}, Wu2020 \cite{wu2020unsupervised}, Jin \cite{jin2022full} and ours. (b0)-(b6) are the corresponding partial enlargement in distorted images of each JND models.}
    \label{details}
    \end{center}
\end{figure*}
\section{Experiments}
\label{Exp}
This section compares the proposed HVS-SD-JND with five anchor methods in terms of details and the JND subjective viewing test. In addition, three ablation studies are performed as well.

\textbf{Datasets:}
Our work is related to two datasets, COCO2017\cite{lin2014microsoft} and CSIQ\cite{larson2010categorical}. Images from COCO2017 datasets are firstly used for training HVS-SD. Then, they are used for training the HVS-SD-JND network. The well-trained HVS-SD-JND is tested on the CSIQ dataset. It should be specifically declared that all the images in the training dataset are cropped into 176 × 176 during training. While the images in the testing dataset are tested under their original size.

\textbf{Anchors:} We compare our proposed method with five previous state-of-the-arts JND methods, which are denoted by Liu2010 \cite{liu2010just}, Wu2013 \cite{wu2013just}, Wu2017 \cite{wu2017enhanced}, Wu2020 \cite{wu2020unsupervised}, and Jin2022\cite{jin2022full}, respectively. The first three methods are HVS-inspired JND models, while the last two are learning-based JND models.

\begin{figure}[htbp]
    \centering
    \includegraphics[width=1.8cm]{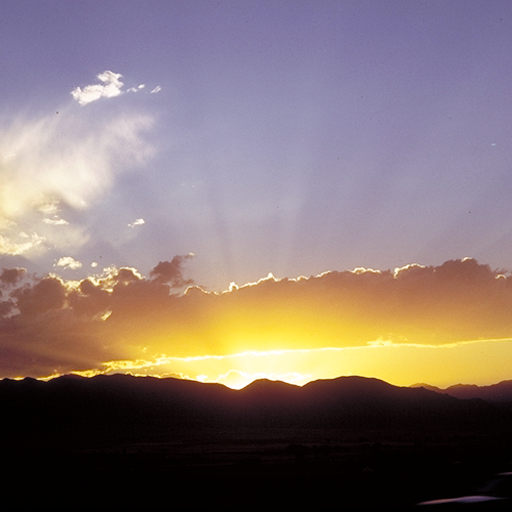}
    \hspace{0.5mm}
    \vspace{0.5mm}
    \includegraphics[width=1.8cm]{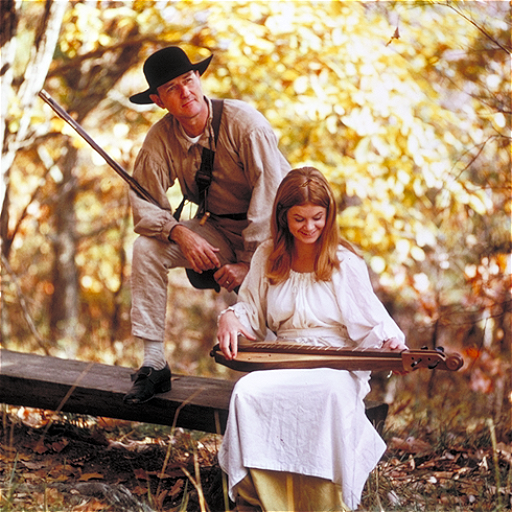}
    \hspace{0.5mm}
    \vspace{0.5mm}
    \includegraphics[width=1.8cm]{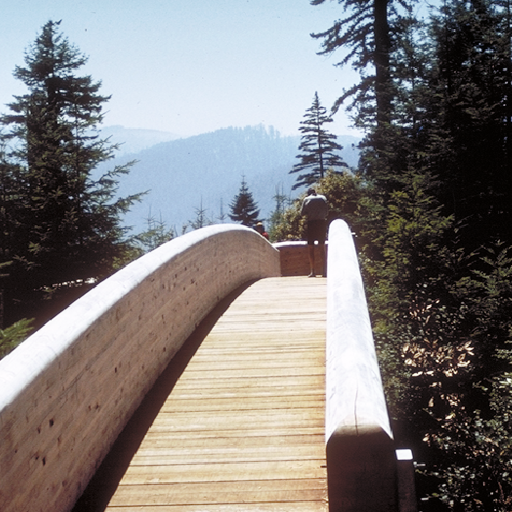}
    \hspace{0.5mm}
    \vspace{0.5mm}
    \includegraphics[width=1.8cm]{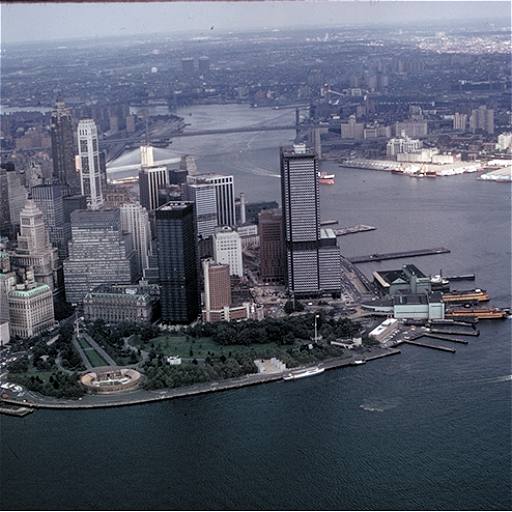}
    \hspace{0.5mm}
    \vspace{0.5mm}
    \includegraphics[width=1.8cm]{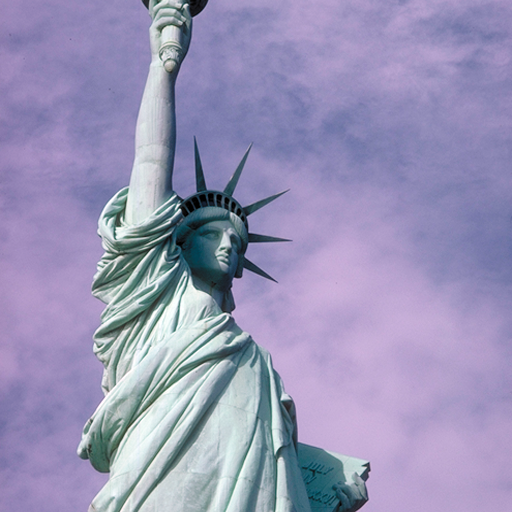}
    \hspace{0.5mm}
    \vspace{0.5mm}
    \includegraphics[width=1.8cm]{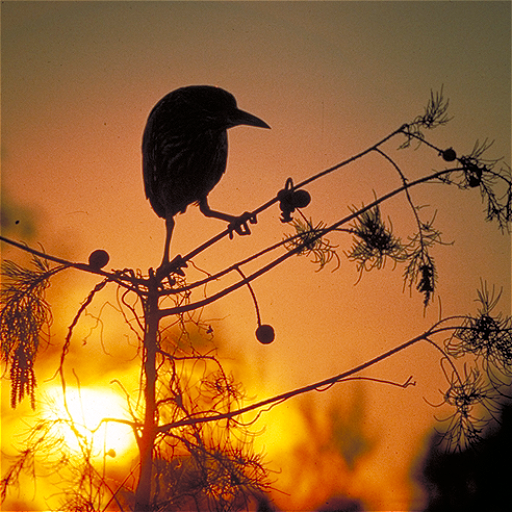}
    \hspace{0.5mm}
    \vspace{0.5mm}
    \includegraphics[width=1.8cm]{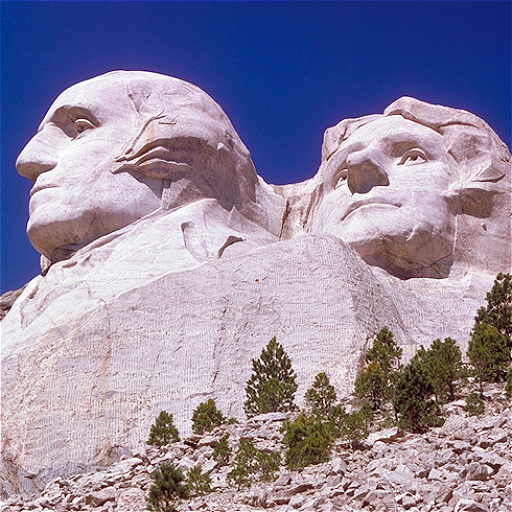}
    \hspace{0.5mm}
    \vspace{0.5mm}
    \includegraphics[width=1.8cm]{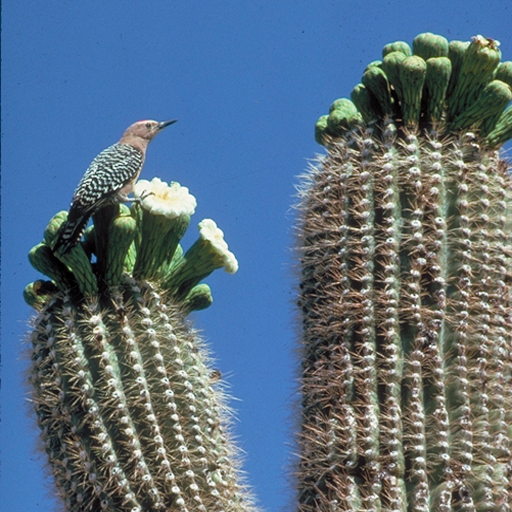}
    \hspace{0.5mm}
    \vspace{0.5mm}
    \includegraphics[width=1.8cm]{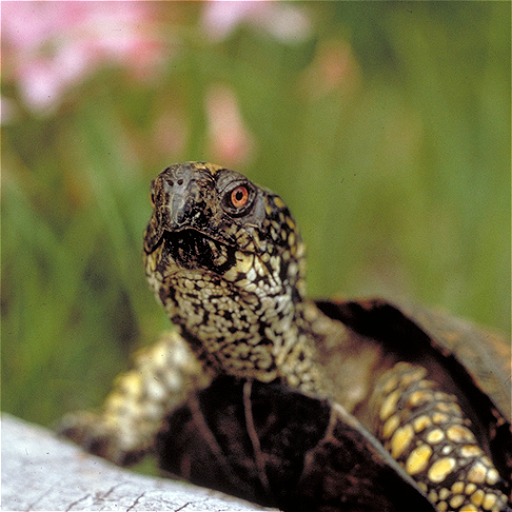}
    \hspace{0.5mm}
    \vspace{0.5mm}
    \includegraphics[width=1.8cm]{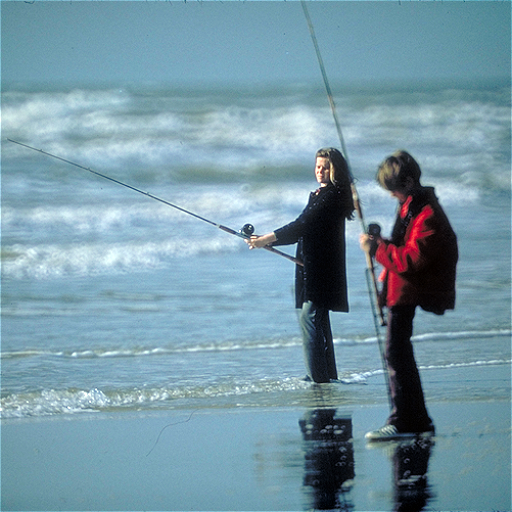}
    \hspace{0.5mm}
    \vspace{0.5mm}
    \includegraphics[width=1.8cm]{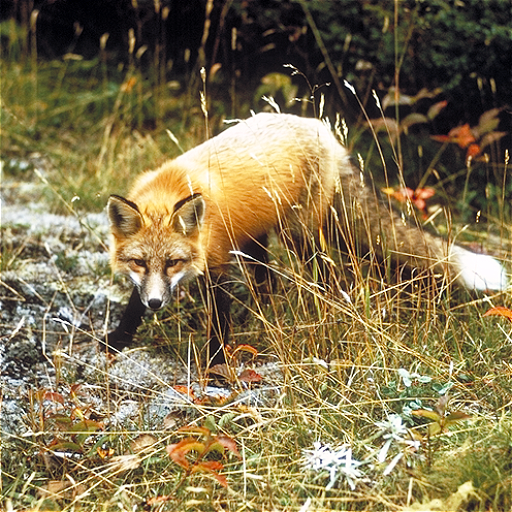}
    \hspace{0.5mm}
    \vspace{0.5mm}
    \includegraphics[width=1.8cm]{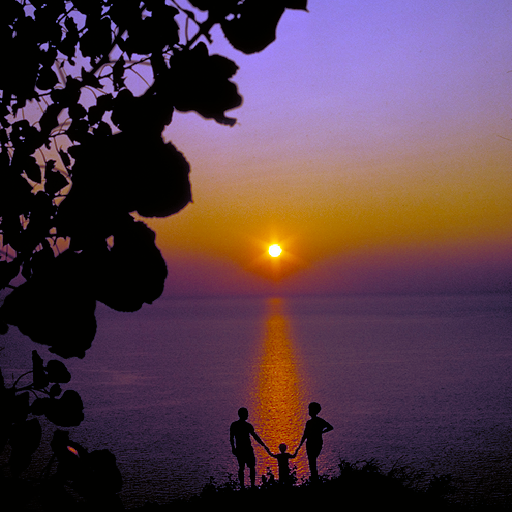}
    \caption{Images from left to right, from top to bottom, are P1-P12 respectively. They are used to carry out JND subjective viewing in our experiments.}
    \label{fig:example}
\vspace{-0.6cm}
\end{figure}

\textbf{Settings:}
JND images $x_j$ are first produced using the aforementioned five anchor methods and our proposed HVS-SD-JND in order to conduct an objective comparison. Then, using $x_j$ as guidance, random noise is injected into the original image $x_0$ to create the JND distorted images $y_0$. This process can be formulated as follows
\begin{equation}
	\begin{split}
	y_0=x_0 \oplus(\varepsilon \cdot r \cdot x_j),
	\end{split}
	\label{formula:adjuster}
\end{equation}
where JND noise level adjuster is denoted by $\varepsilon$. It maintains the level of injected noise from various models (e.g., the same PSNR or MSE). $\oplus$ denotes the addition operation. $r$ represents a random matrix containing positive 1 and negative 1 elements. In the experiments, we inject JND noise as formula \eqref{formula:adjuster}.

During training the HVS-SD-JND, we utilize ADAM\cite{kingma2014adam} optimizer to optimize it. We set the learning rate to $10^{-5}$ during training with batch size 32. We set hyper parameters $\alpha$, $\beta$ and $\gamma$ in formula \eqref{Loss} to 0.1, 1 and 0.1 respectively. Note that the training of the HVS-SD is offline. To ensure that the reconstructed images have high PSNR, we train the HVS-SD with the low QP settings in \cite{balle2016end}.

\subsection{Comparisons with Anchors}
In this subsection, we compare the JND distorted images' details produced by the aforementioned anchor methods and our proposed HVS-SD-JND at first. Then, we perform JND subjective viewing test on these JND distorted images. It should be mentioned that JND subjective viewing test is the most accurate method for evaluating JND models. Therefore, \emph{no IQA metrics} will be used for assessing JND models as they have different drawbacks in their nature as discussed in \cite{jin2022full}.

% \section{Experiments}
% \label{Exp}
% This section compares HVS-SD-JND with one standard baseline, five
% state-of-the-art JND models. In addition, two ablation studies
% are performed. 

\subsubsection{Details comparison}
As shown in Fig. \ref{details}, (a0)-(b0) are the original images. Their associated JND distorted images are exhibited in (a1)-(b1), (a2)-(b2), (a3)-(b3), (a4)-(b4), (a5)-(b5), and (a6)-(b6), which are the images distorted by five anchor models Liu2010 \cite{liu2010just}, Wu2013 \cite{wu2013just}, Wu2017 \cite{wu2017enhanced}, Wu2020 \cite{wu2020unsupervised}, Jin \cite{jin2022full} as well as our proposed HVS-SD-JND with the same level noise at PSNR = 26.06 dB.

As shown in (a1)-(b1) generated with Liu2010\cite{liu2010just}, there is obvious noise in the whole JND distorted image. Such a phenomenon has been improved at some degree in (a2)-(b2) and (a3)-(b3) with Wu2013\cite{wu2013just} and Wu2017 \cite{wu2017enhanced}. However, the region with little gradient still does not perform well in previous HVS-based JND models as (a2)-(b2), (a3)-(b3). All these methods above are HVS-inspired models. (a4)-(b4) shows the results of Wu2020\cite{wu2020unsupervised}, which is an initial work of the learning-based JND model. It tries to generate the JND with the neural networks and assess the reasonability of the generated JND with a single IQA. To better guide the JND generation, pattern complexity is used as a reference input image of the neural network to auto-mining JND thresholds. However, the injected noise in (a4)-(b4) is still obvious. A big promotion is made in (a5)-(b5) generated with Jin2022\cite{jin2022full}. It takes the stimuli of full RGB channels instead of the most sensitive channel into account during JND generation. Besides, the generated JND is assessed with the adaptive IQA combinations (AIC) avoiding the limitation of a single IQA on assessing various distortion types in the signal domain. All these make an obvious improvement in (a5)-(b5) compared with previous anchor methods. However, there is a slight difference that can be perceived from (a5)-(b5), especially for the enlarged regions. While such a slight difference has been further removed in our JND distorted images (a6)-(b6). We firstly propose an HVS-SD network. On the one hand, it represents the signal degradation of the HVS perceptual process in the JND subjective view test and enables the generated JND to be assessed in the perceptual domain, in line with the real world procedure of brain’s visual cortex. On the other hand, it provides more accurate visual attention and sensitivity information for guiding JND generation. Then, we propose a visual attention loss and use it to further control the JND generation. All these advantages make the proposed HVS-SD-JND achieve the SOTA performance in (a6)-(b6). %It's hard to perceive the difference between original image (a0)-(b0) and our JND distorted image (a6)-(b6). 

\begin{table*}[tp]  
	\centering  
	\fontsize{6.9}{8}\selectfont  
	\begin{threeparttable}  
		\caption{\centering{\scshape JND Subjective Viewing Test Results}}
		\label{tab:performance_comparison}  
		\begin{tabular}{ccccccccccc}  
			\toprule  
			\multirow{2}{*}{Index}&  
			\multicolumn{2}{c}{Ours VS. Liu2010\cite{liu2010just}}&
			\multicolumn{2}{c}{Ours VS. Wu2013\cite{wu2013just}}&
			\multicolumn{2}{c}{Ours VS. Wu2017\cite{wu2017enhanced}}&
			\multicolumn{2}{c}{Ours VS. Wu2020\cite{wu2020unsupervised}}&
			\multicolumn{2}{c}{Ours VS. Jin2022\cite{jin2022full}}
            \cr 
		    & Mean & Std & Mean & Std & Mean & Std& Mean & Std& Mean & Std\\
% 			\cmidrule(lr){2-3} \cmidrule(lr){4-5}  
% 			&Precision&Recall&Precision&Recall\cr  
			\midrule  
			P1 & 2.19 & 0.76 & 0.23 & 1.43 & 1.45 & 1.27 & 2.16 & 0.81 & 1.79 &0.97\cr
			P2 & 1.16 & 0.88 & 1.85 & 1.09 & 1.81 & 1.08 & 2.08 & 0.96 & 0.52 &0.96\cr
			P3 & 2.27 & 0.85 & 1.97 & 1.02 & 1.89 & 0.99 & 1.82 & 0.96 & 0.44 & 0.94\cr
        P4 & 2.50 & 0.69 & 2.29 & 0.77 & 2.24 & 0.89 & 2.16 & 0.74 & 1.19 & 1.06\cr
        P5 & 1.85 & 0.96 & 1.08 & 1.47 & 1.15 & 1.38 & 1.50 & 1.17 & -0.10 & 1.25\cr
        P6 & 1.65 & 1.05 & 1.37 & 1.32 & 1.11 & 1.46 & 1.76 & 1.32 & 0.53 & 0.89\cr
        P7 & 2.44 & 0.93 & 1.90 & 1.00 & 1.58 & 1.28 & 2.11 & 0.97 & 1.94 & 0.97\cr
        P8 & 2.08 & 0.97 & 1.69 & 1.21 & 1.66 & 1.18 & 2.02 & 1.13 & 0.98 & 0.99\cr
        P9 & 2.21 & 1.08 & 1.89 & 0.94 & 1.76 & 1.09 & 2.00 & 0.98 & 0.77 & 0.97\cr
        P10 & 1.50 & 1.28 & 1.24 & 1.03 & 0.98 & 1.48 & 1.82 & 1.09 & 0.15 & 1.46\cr
        P11 & 1.42 & 0.98 & 0.65 & 1.36 & 1.00 & 1.23 & 1.13 & 1.33 & 0.26 & 0.93\cr
        P12 & 2.18 & 0.87 & 2.31 & 0.79 & 2.26 & 0.80 & 2.37 & 0.77 & 1.74 & 0.93\cr
% 			A&{\bf 0.8189}&{\bf 0.8139}&{\bf 0.8146}&{\bf 0.6971}\cr  
			\midrule 
			{\bf Average}& {\bf 1.95} & - & {\bf 1.54} & - & {\bf 1.57} & - & {\bf 1.91} & - & {\bf 0.85}\cr
			\bottomrule  
		\end{tabular}  
		\label{tab:mos}
	\end{threeparttable}  
\end{table*}

\subsubsection{JND subjective viewing tests}
\begin{figure}[htbp]
	\begin{center}
		\noindent
		\includegraphics[width = 2.8 in]{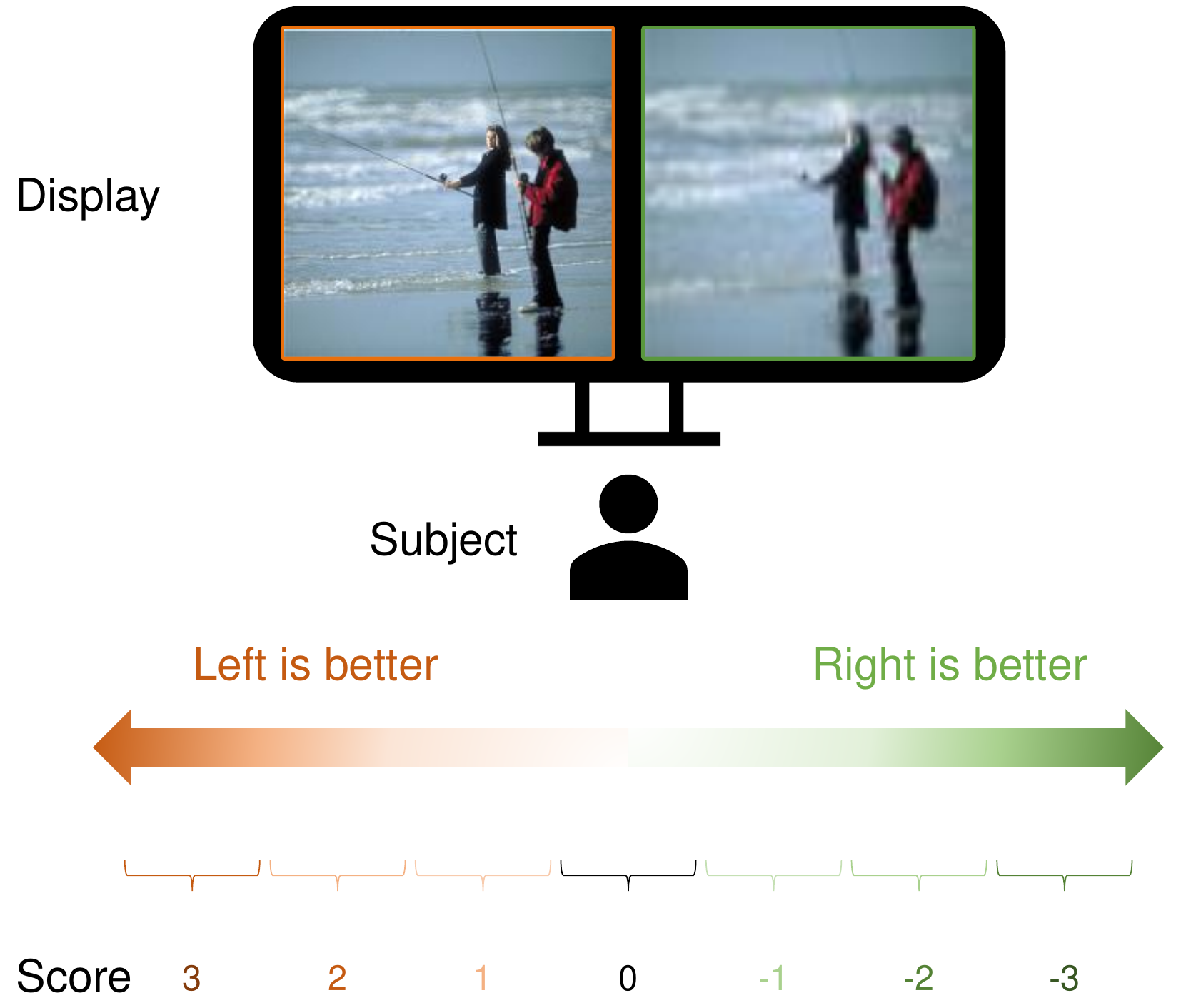}
		\caption{Illustration of the experimental evaluation criteria and scores of subjective test. Scores 3, 2, and 1 represent the left image has much better, better, and a little better quality than the right one. Similarly, scores -3, -2, and -1 represent the right image has much better, better, and a little better quality than the left one. Score 0 represents the left and right images that have a similar quality during the test. }\label{score}
	\end{center}
\end{figure}

As we invite 62 subjects to conduct the JND subjective view test, which is time-consuming and labor-consuming, only the 12 images from the CSIQ dataset that used in \cite{wu2020unsupervised,jin2022full} are selected for the JND subjective viewing test instead of the whole dataset for a fair comparison, as shown in Fig. \ref{fig:example}. The left and right sides of the screen are randomly filled with the distorted images created by our HVS-SD-JND and one of the other anchor methods. None of the test individuals are aware of the connection between the distorted images and the underlying JND models. The evaluation criteria and scores are shown in Fig. \ref{score}. The subject in the experiment was asked to offer a score on images on the screen. The different results of image quality comparison between the two sides are represented with different scores. The score is positive if the left image is better than the right one. Otherwise, the score is non-positive. More details refer to ITU-R BT.500-11 criterion.

% The results of subjective viewing test are shown in TABLE \ref{tab:mos}. We set five anchors, which are Liu2010\cite{}, Wu2013\cite{}, Wu2017\cite{}, Wu2020\cite{} and Jin2022\cite{}. In comparison with Jin2022\cite{}, which is the best JND model recently, we compare image $y_{0}$ and perceptual image $y_{2}$ with this anchor. We involve two metrics into our subjective experiment. One is average scores, denoted by ’Mean’. The other metric is its associated standard deviation, denoted by ’Std’. As all the average scores shown in TABLE \ref{tab:mos}, positive value means that the proposed model is better than the others, and the larger positive value represents the better quality. The standard deviation reflects whether the rating situation for all subjects is unified. Larger standard deviation means larger differences among different subjects during viewing test. Otherwise, there are small differences among subjects.

The JND subjective viewing test results of our proposed method against the other five anchor methods are shown in TABLE \ref{tab:mos}. Two main metrics are involved here. One is average scores, denoted by 'Mean'. The other metric is its associated standard deviation, denoted by 'Std'. For all the average scores shown in TABLE \ref{tab:mos}, the positive value means that the proposed model outperforms the anchor methods, and the larger positive value represents the better performance. The standard deviation shows how consistent the rating conditions are across all subjects. A larger standard deviation means larger differences among different subjects when they are conducting the subjective viewing test. Otherwise, there are small differences among subjects. As shown in TABLE \ref{tab:mos}, the average of the 'Mean' values is positive. Besides, most of the 'Mean' values are larger than 1. These results indicate that our developed model has completely exceeded the other models in terms of the accuracy of the JND estimation. %Moreover, even the perceptual image $y_{2}$ is outperformed than Jin2022\cite{}. 

\subsection{Ablation Studies}
In this subsection, a series of ablation experiments are performed to test the key components of the proposed HVS-SD JND. Similarly, the ablated models are retrained on COCO2017 dataset and tested on the CSIQ dataset during the ablation experiments. All experimental configurations are identical to those described above. Similarly, all the JND distorted images are adjusted to the same PSNR = 26.06 dB in the ablation experiments as well. For comparison, our full model is selected as the baseline here, denoted by $BL$. 
%There are several components of our proposed model. 

\subsubsection{Ablation of the HVS-SD module}
One of the main contributions of this work is the proposed HVS-SD module. It represents signal degradation of the HVS perceptual process in the JND subjective viewing test. With the proposed HVS-SD, the generated JND can be assessed in the perceptual domain.

To evaluate the effects of the proposed HVS-SD, we firstly carry out the ablation of the HVS-SD module. That is, the HVS-SD is removed from the HVS-SD-JND, while the rest parts of the HVS-SD-JND remain. We directly evaluate the image quality in the signal domain by calculating the $Loss2$ on the original image $x_0$ and its associated JND distorted image $y_0$, we have
\begin{equation}
\label{loss2}
Loss2 = AIC(x_0,y_0).
\end{equation}
After that, we retrain the ablated model, termed baseline-perceptual (denoted by $BL-P$). Then, the ablated model is tested on the CSIQ dataset. Finally, we evaluate the performance of $BL$ and $BL-P$ via the JND subjective viewing test. The results are shown in TABLE \ref{tab:mos2}. There is an average of 0.78 quality score improvement with the HVS-SD module. Hence, the HVS-SD matters for the proposed method. The details of images distorted with $BL$ and $BL-P$ are shown in Fig. \ref{fig:ablation} (a0) and (a1). It can be observed that there is more noise in (a1) compared with (a0). Besides, the distortion is uniformly located in the whole image of (a1). It demonstrates that assessing the generated JND in the perceptual domain with the proposed HVS-SD can promote the accuracy of the JND estimation on the whole image. 
\begin{table}[tp]  
	\centering  
	\fontsize{6.9}{8}\selectfont  
	\begin{threeparttable}  
		\caption{\centering{\scshape JND Subjective Viewing Test Results of the Ablation Experiments}}
		\label{tab:performance_comparison}  
		\begin{tabular}{ccccccc}  
			\toprule  
			\multirow{2}{*}{Index}&  
			\multicolumn{2}{c}{$BL$ vs.  $BL-P$}&
			\multicolumn{2}{c}{$BL$ vs. $BL-CAM$}&
			\multicolumn{2}{c}{$BL$ vs. $BL-L3$}
            \cr 
		    & Mean & Std & Mean & Std & Mean & Std\\
% 			\cmidrule(lr){2-3} \cmidrule(lr){4-5}  
% 			&Precision&Recall&Precision&Recall\cr  
			\midrule  
			P1 & 1.39 & 0.89 & 1.60 & 1.02 & 1.69 & 0.98\cr
        P2 & 1.02 & 0.81 & -0.05 & 0.81 & 0.55 & 0.87 \cr
        P3 & 0.05 & 0.75 & 0.83 & 0.79 & 0.61 & 1.11 \cr
        P4 & 1.32 & 0.64 & 1.22 & 0.98 & 1.32 & 1.28 \cr
        P5 & -0.05 & 0.63 & 0.77 & 0.96 & 1.94 & 0.93 \cr
        P6 & 1.03 & 0.74 & 1.08 & 0.98 & 0.00 & 0.95 \cr
        P7 & 1.06 & 0.76 & -0.15 & 0.63 & 0.42 & 0.89 \cr
        P8 & 0.73 & 0.70 & 1.35 & 0.81 & 0.11 & 1.12\cr
        P9 & 1.24 & 0.99 & 0.55 & 0.92 & 0.02 & 0.89\cr
        P10 & 0.13 & 0.55 & 1.20 & 0.93 & 0.18 & 0.81\cr
        P11 & 0.40 & 0.94 & 0.12 & 0.89 & 0.05 & 1.16 \cr
        P12 & 1.05 & 0.61 & 0.62 & 1.13 & 1.27 & 0.81\cr
			\midrule 
			{\bf Average}& {\bf 0.78} & - & {\bf 0.76} & - & {\bf 0.68}&-\cr
			\bottomrule  
		\end{tabular}  
		\label{tab:mos2}
	\end{threeparttable}  
\end{table}

\subsubsection{Ablation of the CAM and guided-CAM}
\begin{figure}[htbp]
    \begin{center}
        \noindent
        \includegraphics[width = 3.3 in]{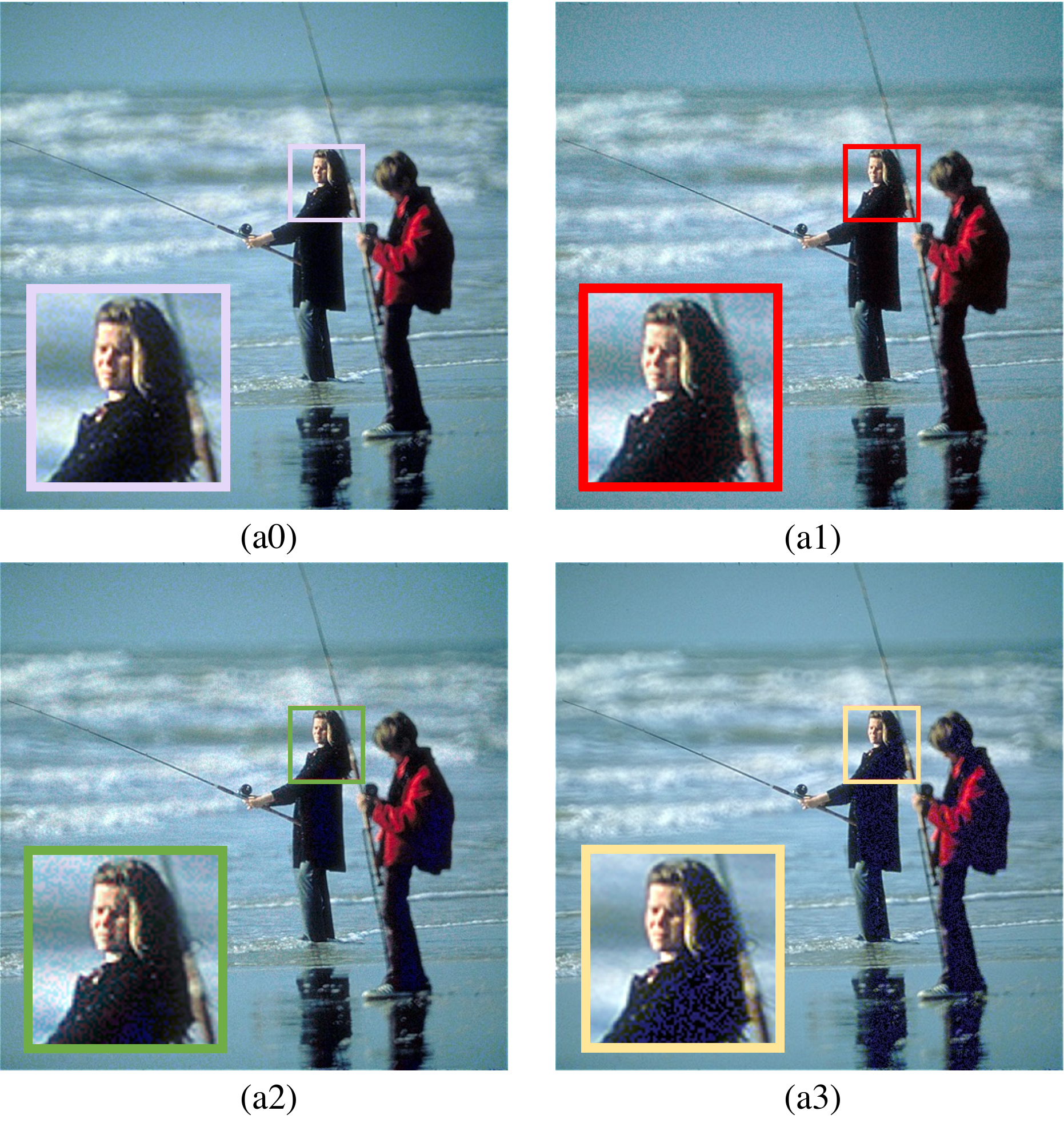}
        \caption{The comparison of degraded models. (a0)-(a3) are $BL$, $BL-P$, $BL-CAM$, and $BL-L3$, respectively. The same regions in these images are enlarged for viewing.}
    \label{fig:ablation}
    \end{center}
\end{figure}

To verify that the CAM and guided-CAM can achieve better performance than the VA and PC used in \cite{jin2022full} during guiding the JND generation, we replace the CAM and guided-CAM with the VA and PC in this subsection. The modified model is termed baseline-CAM, denoted by $BL-CAM$. Similarly, after retraining and testing on COCO2017 and CSIQ, a JND subjective viewing test is performed on the JND distorted images generated with the $BL$ and $BL-CAM$. The results are listed in TABLE \ref{tab:mos2}. There is an average of 0.76 quality score improvement by replacing the VA and PC with the CAM and guided-CAM. Besides, the details of images distorted with $BL-P$ are shown in Fig. \ref{fig:ablation} (a2). It can be observed that obvious color changes appear in (a2), compared with (a0), especially for the dark regions of the foreground. While the distortion in the rest of the regions in (a2) is acceptable. Therefore, compared with the PC and VA, the guided-CAM and CAM can better guide the JND generation, especially for the visual attention focused on foreground regions.

As mentioned in Sec. \ref{sec:introduction}, the VA and PC are handcrafted features, which are obtained based on the HVS-inspired models. Hence, VA and PC have disadvantages by nature due to our limited understanding of the HVS. However, the CAM and guided-CAM represent the attention and sensitivity of the HVS-SD, which are obtained by pre-training on the large dataset. In other words, CAM and guided-CAM can provide more accurate prior information than the handcrafted VA and PC during the JND estimation. Therefore, the CAM and guided-CAM obtained from the pre-trained HVS-SD are more suitable for guiding the JND generation, compared with the handcrafted VA and PC.

\subsubsection{Ablation of the visual attention loss}
In this subsection, we verify the performance of the proposed visual attention loss by removing the $Loss3$ from the total loss, and the other parts of the $BL$ remain the same. The ablated model is termed as the baseline-$Loss3$, denoted by $BL-L3$. Similarly, $BL-L3$ is retrained and tested in the COCO2017 and CSIQ, respectively. After that, JND subjective viewing test is performed on the JND distorted images, distorted with $BL$ and $BL-L3$. The test results are listed in TABLE \ref{tab:mos2}. There is an average of 0.68 quality score improvement by adding visual attention loss $Loss3$ in the total loss. Besides, the details of images distorted with $BL-P$ are shown in Fig. \ref{fig:ablation} (a3). It can be observed that obvious distortion appears in the foreground in (a3), which is the visual attention focused region. While the distortion in the background regions is acceptable. This demonstrates that our proposed visual attention loss can better control the JND generation, especially for the visual attention focused regions.  

The ablation experiments in this section demonstrate that all the key components (i.e., the proposed HVS-SD, CAM, and guided-CAM as well as the proposed visual attention loss) in the proposed HVS-SD-JND network are critical for improving the accuracy of the generated JND.

\section{Conclusion with Discussion}
In this work, we first demonstrate the major drawback of the existing unsupervised-JND generation models: the reasonability of the generated JND is assessed in the signal domain in the real world instead of in the perceptual domain in our
brain. In view of this, we propose an HVS-inspired signal degradation network for just noticeable difference (HVS-SD-JND) estimation. To achieve this, we carefully analyze the HVS perceptual process in the JND subjective viewing test, especially for the signal degradation of the HVS. Then, the signal degradation is represented by the proposed HVS-inspired signal degradation (HVS-SD) network. With the HVS-SD, we can easily infer the degraded image in our brain by inputting the image in the real world. This allows us to assess the reasonability of the generated JND in the perceptual domain. Meanwhile, the HVS-SD can provide more accurate prior information, which is better for guiding JND generation. Considering that reasonable JND will not cause a visual attention shift, we also propose a visual attention loss to further control JND generation. All these advantages in our proposed method above make it achieve the SOTA performance. 

It should be mentioned that the HVS perceptual process is quite complex, and may have different preferences of functions in different tasks. Hence, it still cannot be fully understood due to the limitation of our current knowledge and techniques. However, the signal degradation of the HVS perceptual process in JND subjective viewing tests can be represented with the proposed HVS-SD reasonably. Our ablation experimental results also demonstrate that the HVS-SD can obviously improve the accuracy of the JND estimation for a whole image.

Besides, although image quality assessment (IQA) techniques have been widely studied for decades, there remains a gap between the objective IQA metrics and the HVS perception. That's the main reason that we use the JND subjective viewing test instead of IQA metrics to evaluate the JND distorted images. However, IQA metrics are able to assess the detailed difference between two images. Hence, the adaptive IQA combination loss used in \cite{jin2022full} is inherited for assessing the difference between $x_2$ and $y_2$ in this work.

\ifCLASSOPTIONcaptionsoff
  \newpage
\fi


\begin{thebibliography}{10}
\providecommand{\url}[1]{#1}
\csname url@samestyle\endcsname
\providecommand{\newblock}{\relax}
\providecommand{\bibinfo}[2]{#2}
\providecommand{\BIBentrySTDinterwordspacing}{\spaceskip=0pt\relax}
\providecommand{\BIBentryALTinterwordstretchfactor}{4}
\providecommand{\BIBentryALTinterwordspacing}{\spaceskip=\fontdimen2\font plus
\BIBentryALTinterwordstretchfactor\fontdimen3\font minus
  \fontdimen4\font\relax}
\providecommand{\BIBforeignlanguage}[2]{{%
\expandafter\ifx\csname l@#1\endcsname\relax
\typeout{** WARNING: IEEEtran.bst: No hyphenation pattern has been}%
\typeout{** loaded for the language `#1'. Using the pattern for}%
\typeout{** the default language instead.}%
\else
\language=\csname l@#1\endcsname
\fi
#2}}
\providecommand{\BIBdecl}{\relax}
\BIBdecl

\bibitem{chou1995perceptually}
C.-H. Chou and Y.-C. Li, ``A perceptually tuned subband image coder based on
  the measure of just-noticeable-distortion profile,'' \emph{IEEE Transactions
  on Circuits and Systems for Video Technology}, vol.~5, no.~6, pp. 467--476,
  1995.

\bibitem{yang2005motion}
X.~Yang, W.~Lin, Z.~Lu, E.~Ong, and S.~Yao, ``Motion-compensated residue
  preprocessing in video coding based on just-noticeable-distortion profile,''
  \emph{IEEE Transactions on Circuits and Systems for Video Technology},
  vol.~15, no.~6, pp. 742--752, 2005.

\bibitem{liu2010just}
A.~Liu, W.~Lin, M.~Paul, C.~Deng, and F.~Zhang, ``Just noticeable difference
  for images with decomposition model for separating edge and textured
  regions,'' \emph{IEEE Transactions on Circuits and Systems for Video
  Technology}, vol.~20, no.~11, pp. 1648--1652, 2010.

\bibitem{wu2013just}
J.~Wu, G.~Shi, W.~Lin, A.~Liu, and F.~Qi, ``Just noticeable difference
  estimation for images with free-energy principle,'' \emph{IEEE Transactions
  on Multimedia}, vol.~15, no.~7, pp. 1705--1710, 2013.

\bibitem{wu2017enhanced}
J.~Wu, L.~Li, W.~Dong, G.~Shi, W.~Lin, and C.-C.~J. Kuo, ``Enhanced just
  noticeable difference model for images with pattern complexity,'' \emph{IEEE
  Transactions on Image Processing}, vol.~26, no.~6, pp. 2682--2693, 2017.

\bibitem{jia2006estimating}
Y.~Jia, W.~Lin, and A.~A. Kassim, ``Estimating just-noticeable distortion for
  video,'' \emph{IEEE Transactions on Circuits and Systems for Video
  Technology}, vol.~16, no.~7, pp. 820--829, 2006.

\bibitem{bae2013novel}
S.-H. Bae and M.~Kim, ``A novel dct-based jnd model for luminance adaptation
  effect in dct frequency,'' \emph{IEEE Signal Processing Letters}, vol.~20,
  no.~9, pp. 893--896, 2013.

\bibitem{bae2014novel}
{Bae, Sung-Ho and Kim, Munchurl}, ``A novel generalized dct-based jnd profile
  based on an elaborate cm-jnd model for variable block-sized transforms in
  monochrome images,'' \emph{IEEE Transactions on Image processing}, vol.~23,
  no.~8, pp. 3227--3240, 2014.

\bibitem{niu2013visual}
Y.~Niu, M.~Kyan, L.~Ma, A.~Beghdadi, and S.~Krishnan, ``Visual saliency’s
  modulatory effect on just noticeable distortion profile and its application
  in image watermarking,'' \emph{Signal Processing: Image Communication},
  vol.~28, no.~8, pp. 917--928, 2013.

\bibitem{hadizadeh2017saliency}
H.~Hadizadeh, A.~Rajati, and I.~V. Baji{\'c}, ``Saliency-guided just noticeable
  distortion estimation using the normalized laplacian pyramid,'' \emph{IEEE
  Signal Processing Letters}, vol.~24, no.~8, pp. 1218--1222, 2017.

\bibitem{zeng2019visual}
Z.~Zeng, H.~Zeng, J.~Chen, J.~Zhu, Y.~Zhang, and K.-K. Ma, ``Visual attention
  guided pixel-wise just noticeable difference model,'' \emph{IEEE Access},
  vol.~7, pp. 132\,111--132\,119, 2019.

\bibitem{bae2016dct}
S.-H. Bae and M.~Kim, ``A dct-based total jnd profile for spatiotemporal and
  foveated masking effects,'' \emph{IEEE Transactions on Circuits and Systems
  for Video Technology}, vol.~27, no.~6, pp. 1196--1207, 2016.

\bibitem{jin2016statistical}
L.~Jin, J.~Y. Lin, S.~Hu, H.~Wang, P.~Wang, I.~Katsavounidis, A.~Aaron, and
  C.-C.~J. Kuo, ``Statistical study on perceived jpeg image quality via mcl-jci
  dataset construction and analysis,'' \emph{Electronic Imaging}, vol. 2016,
  no.~13, pp. 1--9, 2016.

\bibitem{wang2016mcl}
H.~Wang, W.~Gan, S.~Hu, J.~Y. Lin, L.~Jin, L.~Song, P.~Wang, I.~Katsavounidis,
  A.~Aaron, and C.-C.~J. Kuo, ``Mcl-jcv: a jnd-based h. 264/avc video quality
  assessment dataset,'' in \emph{IEEE International Conference on Image
  Processing}.\hskip 1em plus 0.5em minus 0.4em\relax IEEE, 2016, pp.
  1509--1513.

\bibitem{wang2017videoset}
H.~Wang, I.~Katsavounidis, J.~Zhou, J.~Park, S.~Lei, X.~Zhou, M.-O. Pun,
  X.~Jin, R.~Wang, X.~Wang \emph{et~al.}, ``Videoset: A large-scale compressed
  video quality dataset based on jnd measurement,'' \emph{Journal of Visual
  Communication and Image Representation}, vol.~46, pp. 292--302, 2017.

\bibitem{liu2018jnd}
X.~Liu, Z.~Chen, X.~Wang, J.~Jiang, and S.~Kowng, ``Jnd-pano: Database for just
  noticeable difference of jpeg compressed panoramic images,'' in \emph{Pacific
  Rim Conference on Multimedia}.\hskip 1em plus 0.5em minus 0.4em\relax
  Springer, 2018, pp. 458--468.

\bibitem{lin2022large}
H.~Lin, G.~Chen, M.~Jenadeleh, V.~Hosu, U.-D. Reips, R.~Hamzaoui, and D.~Saupe,
  ``Large-scale crowdsourced subjective assessment of picturewise just
  noticeable difference,'' \emph{IEEE Transactions on Circuits and Systems for
  Video Technology}, 2022.

\bibitem{liu2019deep}
H.~Liu, Y.~Zhang, H.~Zhang, C.~Fan, S.~Kwong, C.-C.~J. Kuo, and X.~Fan, ``Deep
  learning-based picture-wise just noticeable distortion prediction model for
  image compression,'' \emph{IEEE Transactions on Image Processing}, vol.~29,
  pp. 641--656, 2019.

\bibitem{zhang2021deep}
Y.~Zhang, H.~Liu, Y.~Yang, X.~Fan, S.~Kwong, and C.~J. Kuo, ``Deep learning
  based just noticeable difference and perceptual quality prediction models for
  compressed video,'' \emph{IEEE Transactions on Circuits and Systems for Video
  Technology}, vol.~32, no.~3, pp. 1197--1212, 2021.

\bibitem{tian2021perceptual}
T.~Tian, H.~Wang, S.~Kwong, and C.-C.~J. Kuo, ``Perceptual image compression
  with block-level just noticeable difference prediction,'' \emph{ACM
  Transactions on Multimedia Computing, Communications, and Applications
  (TOMM)}, vol.~16, no.~4, pp. 1--15, 2021.

\bibitem{wu2020unsupervised}
Y.~Wu, W.~Ji, and J.~Wu, ``Unsupervised deep learning for just noticeable
  difference estimation,'' in \emph{IEEE International Conference on Multimedia
  \& Expo Workshops}, 2020, pp. 1--6.

\bibitem{jin2021just}
J.~Jin, X.~Zhang, X.~Fu, H.~Zhang, W.~Lin, J.~Lou, and Y.~Zhao, ``Just
  noticeable difference for deep machine vision,'' \emph{IEEE Transactions on
  Circuits and Systems for Video Technology}, 2021.

\bibitem{jin2022full}
J.~Jin, D.~Yu, W.~Lin, L.~Meng, H.~Wang, and H.~Zhang, ``Full rgb just
  noticeable difference (jnd) modelling,'' \emph{arXiv preprint
  arXiv:2203.00629}, 2022.

\bibitem{hall1977nonlinear}
C.~F. Hall and E.~L. Hall, ``A nonlinear model for the spatial characteristics
  of the human visual system,'' \emph{IEEE Transactions on Systems, Man, and
  Cybernetics}, vol.~7, no.~3, pp. 161--170, 1977.

\bibitem{wu2013perceptual}
H.~R. Wu, A.~R. Reibman, W.~Lin, F.~Pereira, and S.~S. Hemami, ``Perceptual
  visual signal compression and transmission,'' \emph{Proceedings of the IEEE},
  vol. 101, no.~9, pp. 2025--2043, 2013.

\bibitem{kim2015hevc}
J.~Kim, S.-H. Bae, and M.~Kim, ``An hevc-compliant perceptual video coding
  scheme based on jnd models for variable block-sized transform kernels,''
  \emph{IEEE Transactions on Circuits and Systems for Video Technology},
  vol.~25, no.~11, pp. 1786--1800, 2015.

\bibitem{zhang2019divisively}
X.~Zhang, S.~Ma, S.~Wang, J.~Zhang, H.~Sun, and W.~Gao, ``Divisively normalized
  sparse coding: toward perceptual visual signal representation,'' \emph{IEEE
  Transactions on Cybernetics}, vol.~51, no.~8, pp. 4237--4250, 2019.

\bibitem{zhou2020just}
M.~Zhou, X.~Wei, S.~Kwong, W.~Jia, and B.~Fang, ``Just noticeable
  distortion-based perceptual rate control in hevc,'' \emph{IEEE Transactions
  on Image Processing}, vol.~29, pp. 7603--7614, 2020.

\bibitem{nami2022bl}
S.~Nami, F.~Pakdaman, and M.~R. Hashemi, ``Bl-juniper: A cnn-assisted framework
  for perceptual video coding leveraging block-level jnd,'' \emph{IEEE
  Transactions on Multimedia}, 2022.

\bibitem{jia2020rihoop}
J.~Jia, Z.~Gao, K.~Chen, M.~Hu, X.~Min, G.~Zhai, and X.~Yang, ``Rihoop: robust
  invisible hyperlinks in offline and online photographs,'' \emph{IEEE
  Transactions on Cybernetics}, 2020.

\bibitem{cheng2001additive}
Q.~Cheng and T.~S. Huang, ``An additive approach to transform-domain
  information hiding and optimum detection structure,'' \emph{IEEE Transactions
  on Multimedia}, vol.~3, no.~3, pp. 273--284, 2001.

\bibitem{su2020joint}
H.~Su, L.~Yu, and C.~Jung, ``Joint contrast enhancement and noise reduction of
  low light images via jnd transform,'' \emph{IEEE Transactions on Multimedia},
  vol.~24, pp. 17--32, 2020.

\bibitem{gu2015analysis}
K.~Gu, G.~Zhai, W.~Lin, and M.~Liu, ``The analysis of image contrast: From
  quality assessment to automatic enhancement,'' \emph{IEEE Transactions on
  Cybernetics}, vol.~46, no.~1, pp. 284--297, 2015.

\bibitem{shen2013depth}
J.~Shen, D.~Wang, and X.~Li, ``Depth-aware image seam carving,'' \emph{IEEE
  Transactions on Cybernetics}, vol.~43, no.~5, pp. 1453--1461, 2013.

\bibitem{shen2014exposure}
J.~Shen, Y.~Zhao, S.~Yan, X.~Li \emph{et~al.}, ``Exposure fusion using boosting
  laplacian pyramid.'' \emph{IEEE Transactions on Cybernetics}, vol.~44, no.~9,
  pp. 1579--1590, 2014.

\bibitem{zhu2013learning}
G.~Zhu, Q.~Wang, Y.~Yuan, and P.~Yan, ``Learning saliency by mrf and
  differential threshold,'' \emph{IEEE Transactions on Cybernetics}, vol.~43,
  no.~6, pp. 2032--2043, 2013.

\bibitem{cong2022global}
R.~Cong, N.~Yang, C.~Li, H.~Fu, Y.~Zhao, Q.~Huang, and S.~Kwong,
  ``Global-and-local collaborative learning for co-salient object detection,''
  \emph{IEEE Transactions on Cybernetics}, 2022.

\bibitem{sun2022tensorial}
X.~Sun, X.~Zhang, C.~Xu, M.~Xiao, and Y.~Tang, ``Tensorial multiview
  representation for saliency detection via nonconvex approach,'' \emph{IEEE
  Transactions on Cybernetics}, 2022.

\bibitem{selvaraju2017grad}
R.~R. Selvaraju, M.~Cogswell, A.~Das, R.~Vedantam, D.~Parikh, and D.~Batra,
  ``Grad-cam: Visual explanations from deep networks via gradient-based
  localization,'' in \emph{Proceedings of the IEEE international conference on
  computer vision}, 2017, pp. 618--626.

\bibitem{larson2010categorical}
E.~C. Larson and D.~Chandler, ``Categorical image quality (csiq) database,''
  2010.

\bibitem{galkandage2020full}
C.~Galkandage, J.~Calic, S.~Dogan, and J.-Y. Guillemaut, ``Full-reference
  stereoscopic video quality assessment using a motion sensitive hvs model,''
  \emph{IEEE Transactions on Circuits and Systems for Video Technology},
  vol.~31, no.~2, pp. 452--466, 2020.

\bibitem{yu2018perceptually}
L.~Yu, H.~Su, and C.~Jung, ``Perceptually optimized enhancement of contrast and
  color in images,'' \emph{IEEE Access}, vol.~6, pp. 36\,132--36\,142, 2018.

\bibitem{o2002attention}
D.~H. O'Connor, M.~M. Fukui, M.~A. Pinsk, and S.~Kastner, ``Attention modulates
  responses in the human lateral geniculate nucleus,'' \emph{Nature
  neuroscience}, vol.~5, no.~11, pp. 1203--1209, 2002.

\bibitem{tootell1998functional}
R.~B. Tootell, N.~K. Hadjikhani, W.~Vanduffel, A.~K. Liu, J.~D. Mendola, M.~I.
  Sereno, and A.~M. Dale, ``Functional analysis of primary visual cortex (v1)
  in humans,'' \emph{Proceedings of the National Academy of Sciences}, vol.~95,
  no.~3, pp. 811--817, 1998.

\bibitem{ekstrom2008bottom}
L.~B. Ekstrom, P.~R. Roelfsema, J.~T. Arsenault, G.~Bonmassar, and
  W.~Vanduffel, ``Bottom-up dependent gating of frontal signals in early visual
  cortex,'' \emph{Science}, vol. 321, no. 5887, pp. 414--417, 2008.

\bibitem{jonas1992human}
J.~B. Jonas, A.~M. Schmidt, J.~M{\"u}ller-Bergh, U.~Schl{\"o}tzer-Schrehardt,
  and G.~Naumann, ``Human optic nerve fiber count and optic disc size.''
  \emph{Investigative ophthalmology \& visual science}, vol.~33, no.~6, pp.
  2012--2018, 1992.

\bibitem{zrenner2002will}
E.~Zrenner, ``Will retinal implants restore vision?'' \emph{Science}, vol. 295,
  no. 5557, pp. 1022--1025, 2002.

\bibitem{gao2021digital}
W.~Gao, S.~Ma, L.~Duan, Y.~Tian, P.~Xing, Y.~Wang, S.~Wang, H.~Jia, and
  T.~Huang, ``Digital retina: A way to make the city brain more efficient by
  visual coding,'' \emph{IEEE Transactions on Circuits and Systems for Video
  Technology}, vol.~31, no.~11, pp. 4147--4161, 2021.

\bibitem{burr1994selective}
D.~C. Burr, M.~C. Morrone, and J.~Ross, ``Selective suppression of the
  magnocellular visual pathway during saccadic eye movements,'' \emph{Nature},
  vol. 371, no. 6497, pp. 511--513, 1994.

\bibitem{taylor2022representation}
J.~Taylor and Y.~Xu, ``Representation of color, form, and their conjunction
  across the human ventral visual pathway,'' \emph{NeuroImage}, vol. 251, p.
  118941, 2022.

\bibitem{bt2002methodology}
R.~I.-R. BT, ``Methodology for the subjective assessment of the quality of
  television pictures,'' \emph{International Telecommunication Union}, 2002.

\bibitem{dong2011selective}
L.~Dong, W.~Lin, C.~Zhu, and H.~S. Seah, ``Selective rendering with graphical
  saliency model,'' in \emph{2011 IEEE 10th IVMSP Workshop: Perception and
  Visual Signal Analysis}.\hskip 1em plus 0.5em minus 0.4em\relax IEEE, 2011,
  pp. 159--164.

\bibitem{balle2016end}
J.~Ball{\'e}, V.~Laparra, and E.~P. Simoncelli, ``End-to-end optimized image
  compression,'' \emph{arXiv preprint arXiv:1611.01704}, 2016.

\bibitem{abu2019crypto}
M.~Abu-Alhaija, ``Crypto-steganographic lsb-based system for aes-encrypted
  data,'' \emph{International Journal of Advanced Computer Science and
  Applications}, vol.~10, no.~10, 2019.

\bibitem{lin2014microsoft}
T.-Y. Lin, M.~Maire, S.~Belongie, J.~Hays, P.~Perona, D.~Ramanan,
  P.~Doll{\'a}r, and C.~L. Zitnick, ``Microsoft coco: Common objects in
  context,'' in \emph{European Conference on Computer Vision}, 2014, pp.
  740--755.

\bibitem{kingma2014adam}
D.~P. Kingma and J.~Ba, ``Adam: A method for stochastic optimization,''
  \emph{arXiv preprint arXiv:1412.6980}, 2014.

\end{thebibliography}
\end{document}